\documentclass[a4paper,11pt,final]{article}

\usepackage[colorlinks=true]{hyperref}
\usepackage[english]{babel}
\usepackage[utf8]{inputenc}
\usepackage[T1]{fontenc}
\usepackage{lmodern}
\usepackage[pdftex]{graphicx}
\usepackage[top=2cm, bottom=2cm, left=2cm, right=2cm]{geometry}
\usepackage{setspace}
\usepackage[french]{varioref}
\usepackage{lastpage}
\usepackage{fancyhdr}
\usepackage[table]{xcolor}
\usepackage{tikz}
\usepackage{titling}
\usepackage{graphicx}
\usepackage{amsmath}
\usepackage{amsthm}
\usepackage{amssymb}

\usepackage{multirow}
\usepackage{rotating}
\usepackage[font=footnotesize,labelfont=bf]{caption}
\usepackage[gen]{eurosym}
\usepackage{dirtree}

\usepackage{rotating}
\usepackage{colortbl}
\usepackage{subfig}
\usepackage{footmisc}

\usepackage{rotating}
\usepackage{threeparttable, tablefootnote}

\usepackage{diagbox} 

\newtheorem{prblm}{Problem}

\usepackage{tabularx}

\newcolumntype{C}[1]{>{\centering\let\newline\\\arraybackslash\hspace{0pt}}m{#1}}
\newcolumntype{L}[1]{>{\raggedright\let\newline\\\arraybackslash\hspace{0pt}}m{#1}}
\newcolumntype{R}[1]{>{\raggedleft\let\newline\\\arraybackslash\hspace{0pt}}m{#1}}

\usepackage{doi}
\usepackage[autostyle]{csquotes}
\usepackage[style=numeric, citestyle=numeric-comp, 
	maxcitenames=2, maxbibnames=99, 
    backend=biber
]{biblatex}
\usepackage
	{todonotes}

\title{Multiplicity and Diversity: Analyzing the Optimal Solution Space of the Correlation Clustering Problem on Complete Signed Graphs}
\author{Nejat Arinik, Rosa Figueiredo \& Vincent Labatut}

\hypersetup{
    pdftitle={\thetitle},
    pdfauthor={\theauthor},
    bookmarksnumbered=true,bookmarksopen=true,
	unicode=true,colorlinks=true,linktoc=all,
	linkcolor=blue,citecolor=blue,filecolor=blue,urlcolor=blue,
	pdfstartview=FitH
}

\usepackage{enumitem}
\setlist{nolistsep}

\begin{document}
\maketitle
\sloppy

\abstract{In order to study real-world systems, many applied works model them through \textit{signed} graphs, i.e. graphs whose edges are labeled as either \textit{positive} or \textit{negative}. Such a graph is considered as \textit{structurally balanced} when it can be partitioned into a number of modules, such that positive (resp. negative) edges are located inside (resp. in-between) the modules. When it is not the case, authors look for the closest partition to such balance, a problem called \textit{Correlation Clustering} (CC). 
Due to the complexity of the CC problem, the standard approach is to find a single optimal partition and stick to it, even if other optimal or high scoring solutions possibly exist. In this work, we study the space of optimal solutions of the CC problem, on a collection of synthetic complete graphs. We show empirically that under certain conditions, there can be many optimal partitions of a signed graph. Some of these are very different and thus provide distinct perspectives on the system, as illustrated on a small real-world graph. This is an important result, as it implies that one may have to find several, if not all, optimal solutions of the CC problem, in order to properly study the considered system.}


\textcolor{red}{\textbf{Cite as:} N. Arinik, R. Figueiredo \& V. Labatut. Multiplicity and Diversity: Analyzing the Optimal Solution Space of the Correlation Clustering Problem on Complete Signed Graphs, Journal of Complex Networks (forthcoming). DOI: \href{10.1093/comnet/cnaa025}{10.1093/comnet/cnaa025}}

\section{Introduction}

In a signed graph, the edges are labeled as either positive ($+$) or negative ($-$). Such a graph is considered to be balanced, according to the \textit{Structural Balance} theory, if it can be partitioned into two~\cite{Cartwright1956} or more~\cite{Davis1967} modules (i.e. clusters), such that all \textit{positive} (resp. \textit{negative}) edges are located \textit{inside} (resp. \textit{in-between}) these modules. For instance, in a social network whose edges represent like/dislike relationships, this amounts to having mutually hostile social groups of friends.
However, it is very rare for a real-world network to be \textit{perfectly} balanced, in which case one wants to assess the \textit{magnitude} of the imbalance. For a given partition, this imbalance is traditionally measured by counting the number of \textit{frustrated edges}~\cite{Cartwright1956, Zaslavsky1987}, i.e. positive edges located in-between the modules and negative ones located inside them. Computing the graph imbalance amounts to identifying the partition corresponding to the lowest imbalance measure over the space of all possible partitions. This minimization problem is known as the \textit{Correlation Clustering} (CC) problem, proven to be NP-hard~\cite{Bansal2002}.

\begin{figure*}[!htb]
\captionsetup{width=0.9\textwidth}
    \centering
    \subfloat{
        \includegraphics[width=0.28\textwidth]{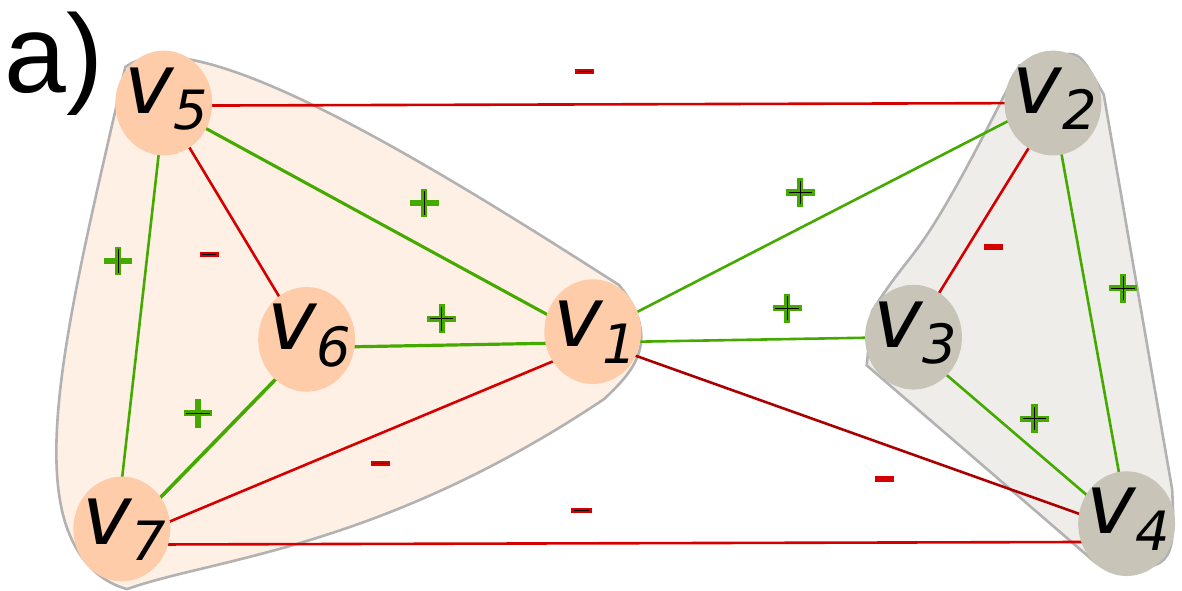}
        \label{fig:intro-ex1a}
    }
    \subfloat{
        \includegraphics[width=0.28\textwidth]{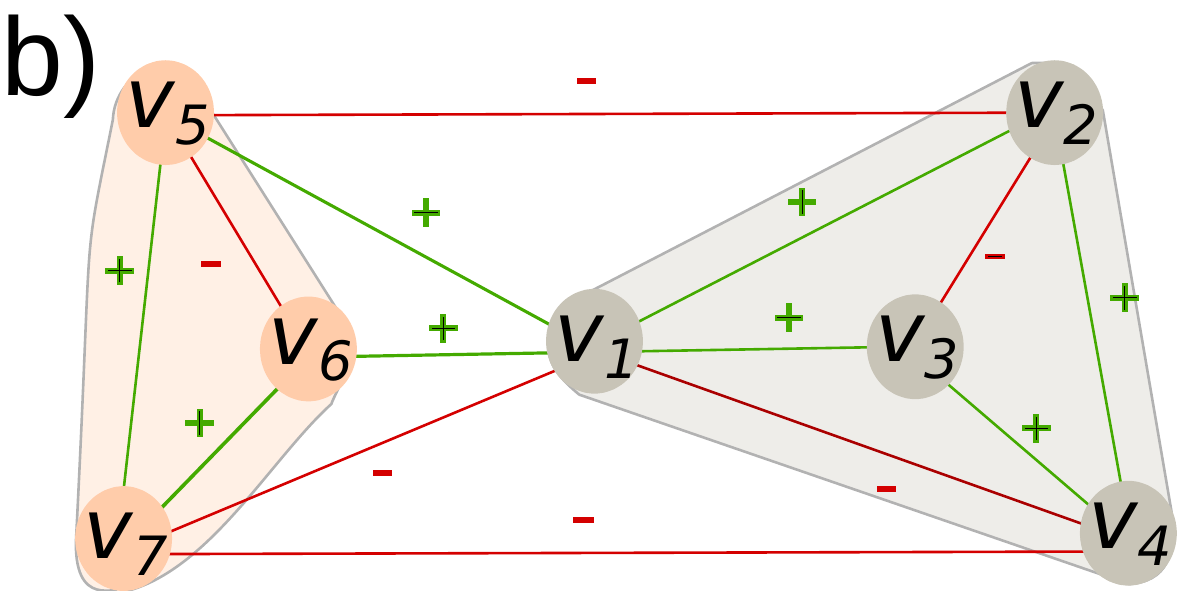} 
        \label{fig:intro-ex1b}
    }
    \subfloat{
        \includegraphics[width=0.28\textwidth]{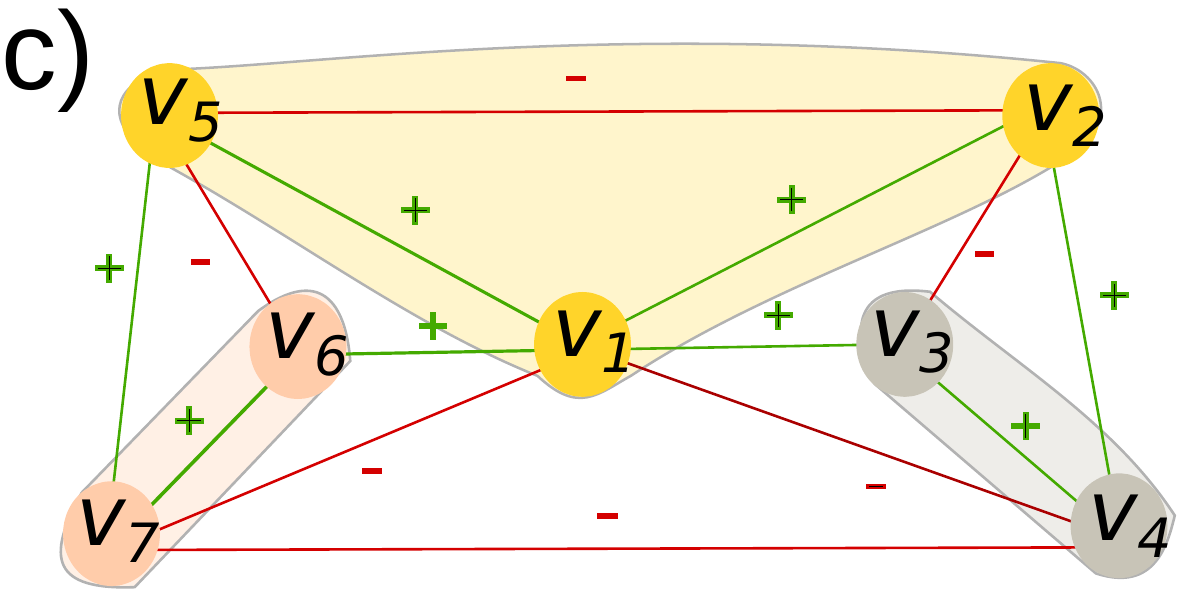}
        \label{fig:intro-ex1c}
    }
    \caption[Fig]{Three (out of 22) different optimal CC solutions obtained for the same network: a) $P_a$ = \{\{$v_1$,$v_5$,$v_6$,$v_7$\}, \{$v_2$,$v_3$,$v_4$\}\}; b) $P_b$ = \{\{$v_5$,$v_6$,$v_7$\}, \{$v_1$,$v_2$,$v_3$,$v_4$\}\}; and c) $P_c$ = \{\{$v_1$,$v_2$,$v_5$\}, \{$v_6$,$v_7$\}, \{$v_3$,$v_4$\}\}.  Red and green lines represent negative and positive edges, respectively. The graph is complete, but for clarity, some negative edges between modules are intentionally omitted. Figure available at \href{https://doi.org/10.6084/m9.figshare.8233340}{10.6084/m9.figshare.8233340}
    under CC-BY license.}
    \label{fig:intro-ex}
\end{figure*}

In the literature, a large number of applied works solve the CC problem to get a better understanding of some studied real-world system, or to solve a specific problem of interest. 
For instance, Jensen~\cite{Jensen2006} constructs a signed spatial network of retail stores to understand the commercial strategies behind their spatial distribution and identify further interesting locations for potential new businesses. The signed relations between vertices encode the spatial interactions between categories of stores: positive edges model attraction (stores of these categories tend to appear in close range) whereas negative ones model repulsion (the opposite). In the domain of time-series analysis, MacMahon \& Garlaschelli~\cite{MacMahon2013} want to capture the structure of different financial markets by studying and comparing the behavior of their traders. For this purpose, they use a signed network whose vertices represent traders and edges model the correlation between the time series of their daily stock returns.
More recently, in~\cite{Arinik2019} we use multiplex signed network to model and study the voting activity of the Members of the European Parliament (MEP). By solving CC, we identify the characteristic ways in which cohesive groups of MEPs form through vote agreement, as well as the legislative contexts leading to such voting patterns.

When solving an instance of the CC problem, most of these works rely on \textit{heuristic} approaches, especially when dealing with large networks, as the problem is NP-hard~\cite{Bansal2002}. But a non-negligible number of studies are also concerned with \textit{optimality}, 
e.g.~\cite{Brusco2010a, Figueiredo2011, Arinik2019, Aref2019}. In any case, the standard approach is to find a \textit{single} solution and focus the rest of the analysis on it, as if it was the \textit{only} solution. Yet, it is possible that several, and even many, other optimal solutions exist for the considered instance, and even more so for \textit{quasi}-optimal solutions. Moreover, these alternate solutions can be very different, in terms of how they partition the graph~\cite{Brusco2010}. 
Figure~\ref{fig:intro-ex} illustrates this on a complete unweighted signed graph (see caption). Solving the CC problem for this graph of only $7$ vertices yields no less than $22$ distinct \textit{optimal} solutions. We show only a few of them to highlight how different these can be. For instance, on the one hand $P_a$ and $P_b$ are very similar, partition-wise, as they are both bisections differing only in the module assignment of $v_1$. On the other hand, $P_c$ is quite different from them: it contains an extra module obtained by separating an element from each module of the previous solutions, in addition to $v_1$. 

Such a focus on a single solution raises several questions. First, as mentioned before, \textit{several} optimal solutions may coexist. If so, one can wonder which network properties lead to this situation, and how many of these solutions are equally relevant to the application problem at hand. Perhaps it would be necessary to design a more appropriate version of the CC problem, in order to distinguish them, possibly based on some additional criteria related to the application context. Second, how different are these solutions? Application-wise, very similar solutions 
could be given the same interpretation, 
whereas substantially different ones might correspond to dramatically different ways of seeing the studied system.
Third, when dissimilar solutions coexist for the same problem, is it possible to detect classes of similar solutions? Indeed, if such classes exist, one could need only to find one representative solution in each class, which would ease the exploration of the solution space. Fourth and finally, in case of the existence of multiple such classes, what distinguishes them from each other? Identifying these characteristic differences could provide some valuable information to understand the studied system. More generally, the answers to all these questions can drive the choice of the method used to solve CC.
 
In this work, our goal is to answer these questions through the characterization of the space of optimal solutions associated with a collection of signed graphs. We proceed by randomly generating a number of signed networks with various characteristics: number of vertices, number of modules, and imbalance. We then identify all their optimal solutions, and study how the number and nature of these solutions is affected by the network characteristics. 
Finding all optimal solutions is computationally costly, and constrains the number of vertices that we can handle. We could instead look for \textit{quasi}-optimal solutions, which can be found faster using a heuristic method. However, we want to study the CC problem \textit{itself}, and not some of its existing resolution methods. Using a heuristic-based method would introduce a bias in the way the solution space is explored, and thus in our study of the problem.

As a first step of a longer-term work, in this article we focus on \textit{unweighted complete} signed networks, which constitute the simplest form of signed graphs. 
They are far from being as popular as sparse graphs when it comes to representing real-world systems, and are mainly restricted to certain types of applications (e.g. vote networks~\cite{Kropivnik1996, Arinik2019}, see Section~\ref{subsec:network-generation} for other examples). However we deem them more appropriate, in the context of this first work, because they allow us to study the problem space while focusing on the most \textit{essential} parameters. We therefore leave other types of graphs (incomplete and/or weighted) to future work, hence postponing the treatment of important (but nevertheless secondary) issues such as the effect of density and degree distribution on the solution space.
Our main contribution is essentially twofold. First, we propose a method to study the solution space, which is generic enough to be applied to other combinatorial problems. 
Second, we obtain exciting results for the CC problem, which open some very interesting perspectives regarding the improvement of resolution methods and their use on specific real-world applications. Finally, as a minor contribution we also propose an open source random model to generate complete unweighted signed graphs with controlled balance.

The rest of the article is organized as follows. In Section~\ref{sec:RelatedWork}, we review the works related to the enumeration and analysis of solution spaces containing multiple optimal solutions. In Section~\ref{sec:CCproblem}, we then give the formal definition of the CC problem, and justify our approach through a few simple examples. We turn to the methods in Section~\ref{sec:Methods}, and explain the approach we propose for the analysis of the space of optimal solutions for CC. In Section~\ref{sec:Experiments}, we present our random model as well as the  synthetic signed networks generated to conduct our analysis. In Section~\ref{sec:Results}, we describe and discuss our results, in order to answer the questions asked above. Finally, in Section~\ref{sec:Conclusion} we summarize our findings, comment the limitations of our work and describe how they could be overcome, and how our work can be extended.


\section{Related Work}
\label{sec:RelatedWork}


As we explain in Section~\ref{subsec:ExactCC}, there is only a very limited number of methods proposed in the literature to solve CC \textit{exactly}. Some of them, as well as subsequent works, identify the issue of multiple optimal solutions, but only scratch the surface, as summarized in Section~\ref{subsec:ExistMultOptSol}. Indeed, exploring the space of optimal solutions requires to deal with additional methodological points, in particular: getting the complete set of optimal solutions for the considered instance, and determining how similar or different they are. However, we could not find any work dealing with this for CC in the literature. For this reason, we widen the scope of our review on these aspects, and consider works conducted on other problems than CC. In Section~\ref{subsec:ComparSol}, we focus on the comparison of solutions; in Section~\ref{subsec:EnumOptSol} we present the main methods used to enumerate optimal solutions; and finally in Section~\ref{subsec:DiversitySol} we review works concerned with the diversity of these solutions.

\subsection{Exact Resolution of CC}
\label{subsec:ExactCC}
In the literature, we find two works solving \textit{exactly} the CC problem by using two different optimization methods: an \textit{ad hoc} combinatorial Branch-and-bound (B\&B) programming method~\cite{Brusco2010a} and one based on Integer Linear Programming (ILP)~\cite{Demaine2006}. Both essentially rely on B\&B~\cite{Land1960}.
Despite its genericity, the ILP approach remains more powerful than the former, because it is based on mathematical modeling (e.g. linear relaxation, dual tightening). Indeed, Figueiredo \& Moura~\cite{Figueiredo2013b} performed a computational experiment with both approaches for the CC problem, and showed that ILP (see Appendix~\ref{apx:Model} for their model formulation) can handle larger graphs (in terms of number of vertices) and in most cases performs better in terms of running time.
Some works also deal with the exact solution of some variants of the CC problem,~\cite{Brusco2010,Brusco2010a,Figueiredo2013b,Aref2019}.

In any case, the primary concern of all these exact optimization methods is only to find a single optimal solution, 
and they ignore or overlook the issue of multiplicity.

\subsection{Existence of Multiple Optimal Solutions}
\label{subsec:ExistMultOptSol}

To the best of our knowledge, the issue of multiple optimal solutions for the CC problem is first pointed out by Davis~\cite{Davis1967} for perfectly balanced \textit{incomplete} signed graphs, as he gives an example of how such graphs may have several optimal partitions. 
In particular, he states that a signed graph should have a unique optimal partition, otherwise, it amounts to a lack of cluster structure. The issue is then also confirmed by Doreian and Mrvar~\cite{Doreian1996} (also in \cite{Doreian2005} with more networks) for imbalanced \textit{incomplete} signed graphs, and the authors integrate this knowledge into their heuristic method to collect all discovered best partitions across a large number of restarts, evidently with the risk of including local optima. Later, Brusco \textit{et al.}~\cite{Brusco2010a} overcome this local optima issue by adapting their \textit{ad hoc} B\&B programming method to enumerate multiple optimal partitions with a limit up to $2,000$ partitions. 

Although considerable efforts are made in both of these works to deal with the multiplicity of solutions, their authors do not try to study the optimal solution space of the CC problem. This might be due to the fact that the number of optimal partitions they encountered was small, around 20, for most of the networks they considered~\cite{Doreian2005}. 
Doreian \textit{et al.}~\cite{Doreian2005} suggest to use the multiplicity as an additional criteria to select the most appropriate number of modules, in cases where the optimal imbalance value is reached for several values of this number of modules. 
Brusco \textit{et al.} apply this principle in \cite{Brusco2010a}. However, as we will show in Section~\ref{sec:Results}, in practice the number of optimal solutions can be much larger than 20.

The problem of multiplicity is of general interest, and was studied in the context of other optimization problems than CC. There are just a few of these works, thus for the sake of completeness we briefly cover them here. Paris~\cite{Paris1981} proposes to take advantage of multiple optimal solutions to perform a more thorough validation of linear programming economic models, in order to provide more flexibility at decision-making.
Liu \textit{et al.}~\cite{Liu2012} tackle the multiplicity for the \textit{Optimal Load Distribution} problem to manage multiple generator units in hydropower plants.
Ruiter~\textit{et al.}~\cite{Ruiter2016} show in the context of Adjustable Robust Optimization that even when all optimal solutions have the same worst-case cost, their mean costs can drastically differ, which allows discriminating between optimal solutions.
Arthur \textit{et al.}~\cite{Arthur1997} also recognize the need to identify all optimal solutions for the \textit{Maximal Covering} problem in the context of geosciences. In addition, they observe a connection between the size of the problem (number of units) and the number of optimal solutions.

All these works show that 1) there can be multiple optimal solutions in practical contexts; and 2) identifying all or several of these multiple solutions is informative, and therefore worthwhile, as they can be leveraged to improve the results application-wise. 
Among other things, the work we present in this article extends the findings of Davis~\cite{Davis1967} and Doreian \textit{et al.}~\cite{Doreian2005} by showing that the issue of multiplicity also occurs for \textit{complete} imbalanced signed graphs. Moreover, we study how certain parameters of the problem affect the multiplicity of solutions.

\subsection{Comparison Between Solutions}
\label{subsec:ComparSol}
The works from the previous paragraph identify the existence and relevance of multiple optimal solutions, but do not try to compare them, or only in terms of cost~\cite{Doreian1996, Arthur1997}. From this perspective, the approach of Good \textit{et al.}~\cite{Good2010} is interesting, even if it deals with \textit{sub-optimal} solutions, as it aims to compare the \textit{nature} of these solutions. They study the \textit{Modularity Maximization} problem, which consists in detecting a community structure in an unsigned graph, i.e. to partition it in order to get cohesive and well separated modules. They show that this problem admits an exponential number of distinct quasi-optimal solutions, and that moreover, these can be structurally very different, an issue they call \textit{degeneracy}.

The connection with our own work is double. First, they deal with the partitioning of graphs, albeit unsigned ones. Second, we perform a similar comparison between graph partitions, with the difference that we focus only on \textit{optimal} solutions. Such comparison is very important, as one can consider a given solution as a view or interpretation of the studied system. Therefore, in addition to identifying the multiplicity of optimal solutions, it is necessary to study \textit{how much} they differ.

\subsection{Enumeration of Optimal Solutions}
\label{subsec:EnumOptSol}
In order to study the optimal solutions, it is necessary to enumerate them, without producing the same solution twice. In our case, the literature provides only two main methods to do so efficiently: \textit{B\&B} vs. \textit{Parameterized Enumeration}.

For B\&B, the enumeration of all optimal solutions has been performed using three different algorithmic approaches. The first is based on ILP and relies on an iterative process: the original problem is solved as many times as there are optimal solutions, as in~\cite{Arthur1997}. This is practically possible when already-found optimal solutions are iteratively added as constraints into the mathematical model to exclude them. The drawback is that this requires building a branch-and-bound search tree from scratch at each iteration. Brusco \textit{et al.}~\cite{Brusco2010a} reduce the number of iterations with a two-step \textit{ad hoc} combinatorial B\&B programming method. They first identify an optimal solution, allowing them to get the optimal objective function value. They then use this value as input when building the branch-and-bound trees corresponding to the remaining solutions. However, these are built from scratch, without leveraging the first tree. Moreover, the process is repeated for each considered number of modules.
Danna \textit{et al.}~\cite{Danna2007} propose a more efficient two-step method, as they not only enumerate all the optimal solutions based on the search tree of the first step, but also take advantage of mathematical modeling to apply efficient search techniques (e.g. dual tightening).
This method is incorporated in the industrial optimization solver CPLEX~\cite{cplex12}.

The parameterized enumeration approach is valid only if the considered problem is controlled by some parameter. Damaschke~\cite{Damaschke2010} proposes an FPT (Fixed-Parameter Tractable) algorithm to enumerate all optimal solutions in a given graph for the \textit{Cluster Editing} problem, which is equivalent to the CC problem when the input graph is complete and unweighted. A drawback about this FPT algorithm is its problem-dependency, i.e. one cannot simply use the same enumeration algorithm to handle other types of networks (e.g. incomplete networks), as opposed to ILP and B\&B programming. Furthermore, Damaschke does not provide the computational results in his theoretical work.

\subsection{Diversity of Solutions}
\label{subsec:DiversitySol}
Enumerating all optimal solutions is costly, so one alternative is to discover only \textit{certain} of them, often with some additional criterion of diversity
(similar in principle to multi-objective optimization approaches). Appa~\cite{Appa2002} proposes an LP-based algorithm which, 
starting from an already-found optimal solution, finds an alternative optimal solution which is as different as possible. One can apply the method a number of times to sample the space of optimal solutions, and then select the most diverse ones. Danna \textit{et al.}~\cite{Danna2007} do the same but for binary linear models. In the context of \textit{Data Clustering}, some methods such as~\cite{Jain2008} have been proposed to detect multiple partitions, but these are not necessarily optimal. 

In any case, the limitation of these methods is usually the estimation of the correct number of solutions: if it is underestimated, the solution space may not be sufficiently covered, whereas if it is overestimated, the computational cost stays high. 
%
Here, the connection with our work is the idea that diversity is important when dealing with multiple optimal solutions. We explore this aspect through the notion of \textit{classes of similar solutions}, which we define later in Section~\ref{sec:Methods}.


\section{Correlation Clustering Problem}
\label{sec:CCproblem}
In this section, we give the mathematical formulation of the CC problem (Section~\ref{subsec:CCformula}), before showing with examples how structurally very similar or very different solutions can be formed  (Section~\ref{subsec:BackgroundMultiplicity}).

\subsection{Mathematical Formulation}
\label{subsec:CCformula}
Let us first introduce our notations before defining CC. Let $G=(V,E)$ be an \textit{undirected graph}, where $V$ and $E$ are the sets of vertices and edges, respectively. We note $n=|V|$ and $m=|E|$ the numbers of vertices (i.e. network order) and edges, respectively. Consider a function $s : E \rightarrow \{+,- \}$ that assigns a \textit{sign} to each edge in $E$. An undirected graph $G$ together with a function $s$ is called a \textit{signed graph}, denoted by $G = (V, E, s)$. An edge $e \in E$ is called negative if $s(e) = -$ and positive if $s(e) = +$. We note $E^-$ and $E^+$ the sets of negative and positive edges,  respectively. 

Let $P = \{M_1,...,M_\ell\}$ ($1 \leq \ell \leq n$) be an $\ell$-partition of $V$, i.e. a division of $V$ into $\ell$ non-overlapping and non-empty subsets $M_i$ ($1 \leq i \leq \ell$) called \textit{modules}. The partition $P$ is called a \textit{solution} for the given graph. Given a solution $P$, an edge is called \textit{internal} if it is located inside a module, or \textit{external} if it is located between any two modules. 
For $\sigma \in \{+,-\}$, the total number of positive or negative edges (depending on $\sigma$) connecting two modules $M_i, M_j \in P$ ($1 \leq i,j \leq \ell$) is noted $\Omega^\sigma(M_i,M_j)$.

The \textit{Imbalance} $I(P)$ of a solution $P$ is defined as the number of \textit{frustrated} edges, i.e. the total number of positive edges located between modules and of negative edges located inside them, i.e.

\begin{equation}
	I(P) = \sum_{1\leq i\leq \ell} \Omega^-(M_i,M_i) +  \sum_{1\leq i<j\leq \ell} 	\Omega^+(M_i,M_j).
	\label{IP}
\end{equation}

The Correlation Clustering problem is formally described as follows:

\begin{prblm}[CC problem]
	For an unweighted signed graph $G=(V,E,s)$, the \textit{Correlation Clustering problem} consists in finding a partition $P$ of $V$ such that the imbalance $I(P)$ is minimized.
\end{prblm}

To the best of our knowledge, this $NP$-hard minimization problem appears under this name for the first time in Bansal's paper~\cite{Bansal2002}. Nevertheless, it was formalized before in the literature, e.g. in~\cite{Doreian1996}, where a local optimization method was also presented.

\subsection{Illustrative Cases}
\label{subsec:BackgroundMultiplicity}
\begin{figure*}[!htb]
\captionsetup{width=0.9\textwidth}
    \centering
    \subfloat{
        \includegraphics[width=0.3\textwidth]{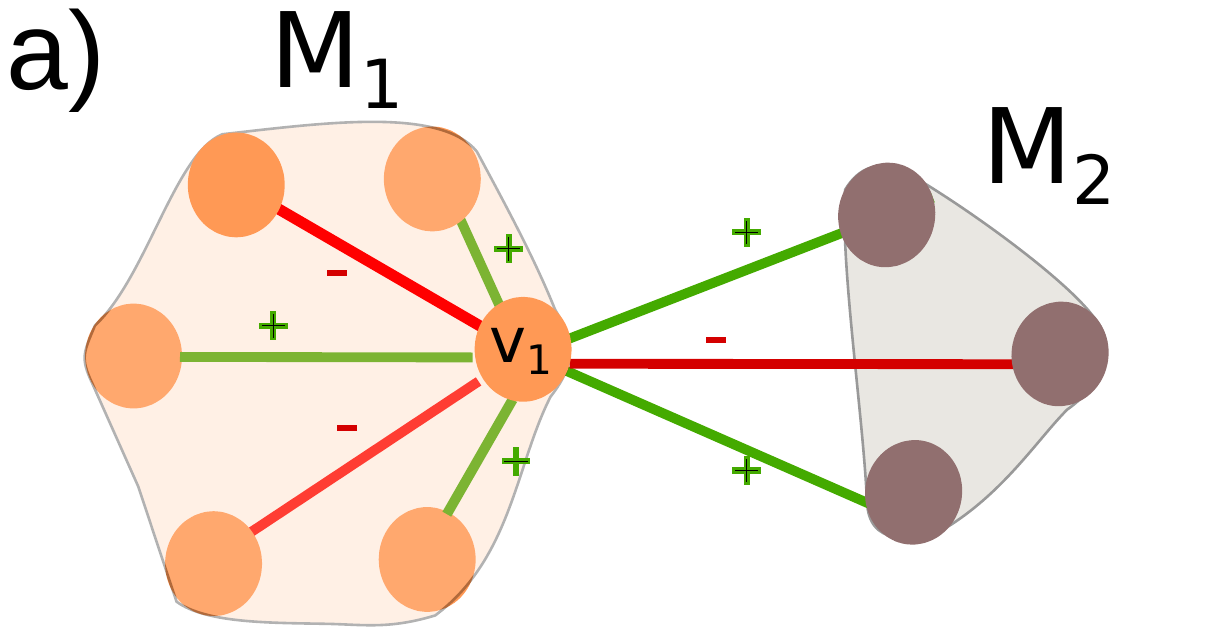}
        \label{fig:CC-ex1a}
    }
    \hspace{1cm} 
    \subfloat{
        \includegraphics[width=0.3\textwidth]{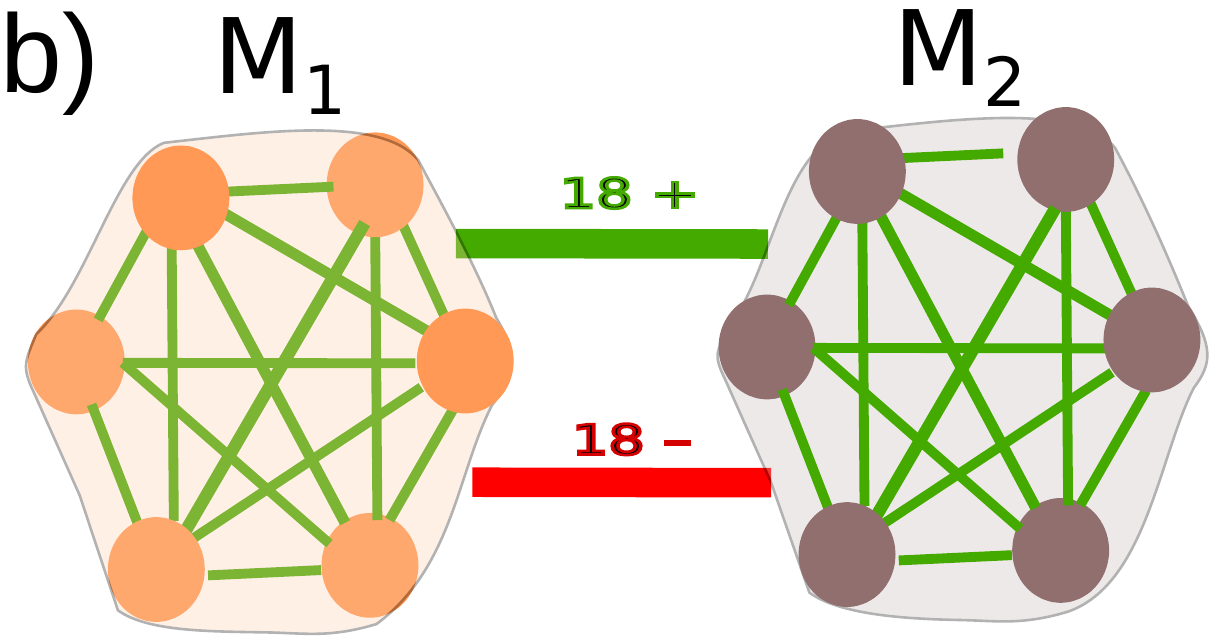}
        \label{fig:CC-ex1b}
    }
    \caption[Fig]{Illustrative examples regarding optimal multiplicity for the CC problem. 
    Figure available at \href{https://doi.org/10.6084/m9.figshare.8233340}{10.6084/m9.figshare.8233340} under CC-BY license.}
    \label{fig:CC-ex1}
\end{figure*}

Figure~\ref{fig:CC-ex1a} gives an example of the kind of situation that can lead to two very similar optimal solutions in complete signed graphs. Other examples exist in the literature for \textit{incomplete} signed graphs, e.g.~\cite{Davis1967, Doreian2005}. Note that the graph in Figure~\ref{fig:CC-ex1a} is fully connected, but only the edges attached to $v_1$ are represented, for matters of readability. The displayed bisection (i.e. modules $M_{1}$ and $M_{2}$) corresponds to an optimal solution. Consequently, the module assignment of $v_{1}$ is optimal as well. This implies that the signed sum of its \textit{external} edges towards any module (currently, $+1$ for $M_{2}$) cannot be greater than that of its \textit{internal} edges (currently, $+1$ for $M_{1}$). This case of equality between the internal and external edges means that moving $v_{1}$ to module $M_2$ instead of $M_1$ does not change the imbalance. Consequently, this change produces another optimal solution.

This example shows how two similar optimal solutions can be obtained through a simple vertex change (see also Figures~\ref{fig:intro-ex1a} and~\ref{fig:intro-ex1b}). Of course, the same principle can be extended to larger changes involving more vertices (e.g. Figure~\ref{fig:intro-ex1c}). Our point here is that it is relatively straightforward to explain the existence of multiple similar optimal solutions. However, it is equally easy to give examples of very different optimal solutions, as well.
%
For instance, Figure~\ref{fig:CC-ex1b} shows the case of a network constituted of two positive cliques, both connected by the same number of positive and negative edges. Again, the network is complete, but only the relevant edges are displayed. Solving the CC problem for this network yields a bisection whose modules $M_{1}$ and $M_{2}$ correspond to the positive cliques. But putting all the vertices in the same module is also an optimal solution: in both cases, the imbalance is 18. This example shows that structurally different optimal solutions can coexist for the CC problem. 

In conclusion, we have shown that it is possible to obtain structurally very similar as well as very different optimal solutions when solving the CC problem. However, we do not know whether these situations coexist in the same solution space, how frequent they are, or how this depends on the graph topology. These observations motivate us to adopt a more systematic approach for further investigations in the rest of this article.

\section{Methods}
\label{sec:Methods}
In this section, we describe the method that we propose to analyze the space of optimal solutions for the CC problem. First, we need to clarify our terminology, as we handle various types of partitions. 
As mentioned before, the \textit{optimal solution} (or \textit{solution} for short) obtained by solving CC for a given graph is a partition of the vertex set that minimizes the imbalance measure. A subset of vertices in this partition is called a \textit{module}. We reserve the term \textit{clustering} to refer to a partition of the set of all solutions. A subset of solutions in such clustering is simply called a \textit{solution class} (or a \textit{class}, for short).

\begin{figure}[!ht]
\captionsetup{width=0.9\textwidth}
\centering
\includegraphics[width=0.9\textwidth,clip=true]{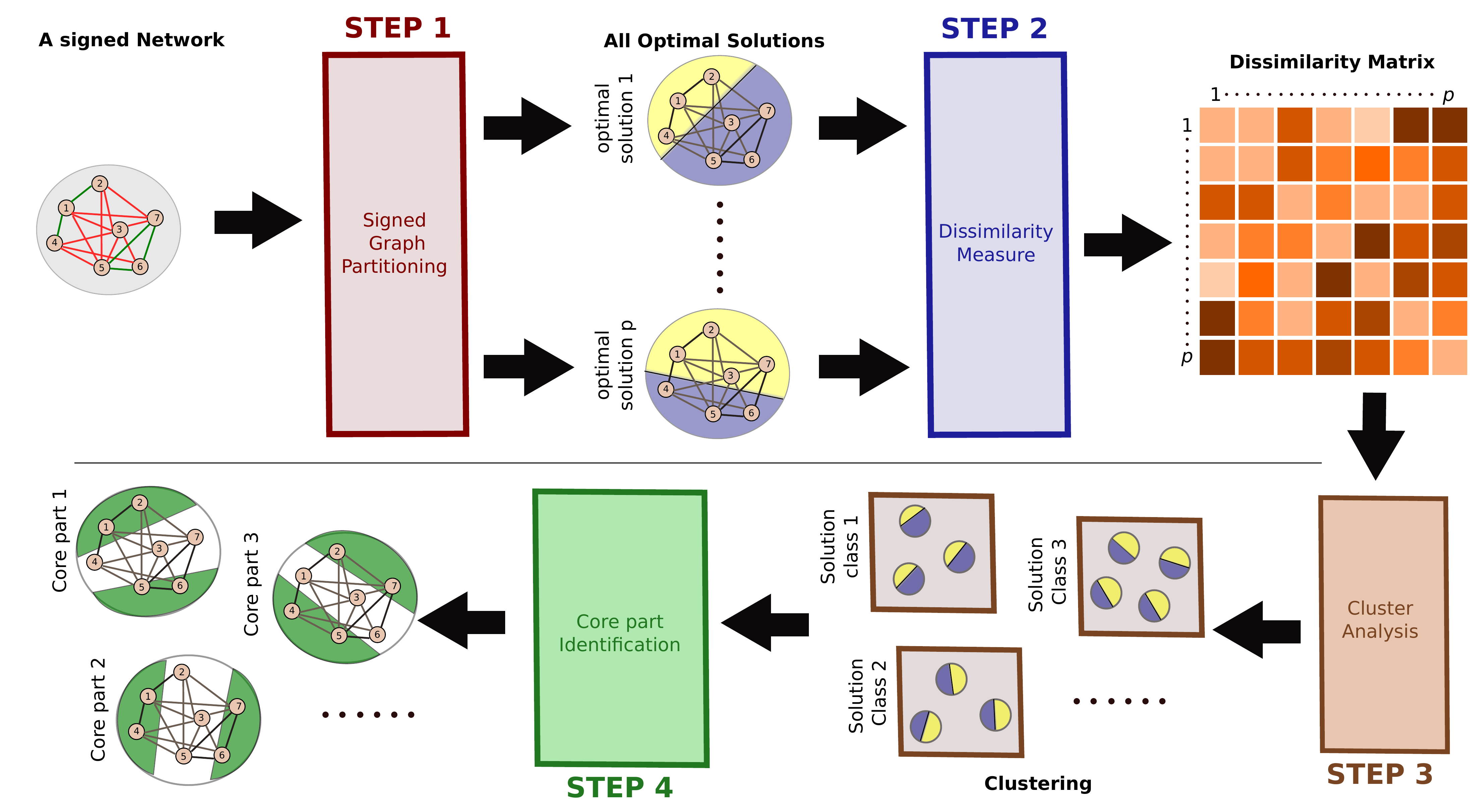}
\caption{Workflow proposed to study the solution space.
Figure available at \href{https://doi.org/10.6084/m9.figshare.8233340}{10.6084/m9.figshare.8233340} under CC-BY license.}
\label{fig:workflow}
\end{figure}


Our goal here is to determine whether it is worth enumerating all optimal solutions when solving CC for a given application. Put differently, we want to know what we lose when we consider only one solution, while there might be multiple ones. To this aim, we propose a $4$-step pipeline approach which is represented in Figure~\ref{fig:workflow}. Each step allows answering a question that naturally arises in our analysis of the space of optimal solutions, and it is implemented through a well-known existing tool deemed appropriate for this purpose. Our methodological contribution is found in the combination of these tools to build our pipeline. 
The input of the pipeline is a complete unweighted signed network. The first step is to enumerate all optimal solutions for this network (Section~\ref{subsec:EnumOptPartitions}), allowing to determine whether \textit{several optimal solutions coexist}. If so, the second step consists in computing the dissimilarity between them (Section~\ref{subsec:ComputingDistance}), in order to assess \textit{how different the obtained solutions are}. The third step consists in performing a cluster analysis of the solutions (Section~\ref{subsec:PerformingClustering}), to check for \textit{the existence of classes of similar solutions}. 
If there are several of them, the fourth and final step is to identify their \textit{core parts} (Section~\ref{subsec:CorePartIdentification}), in order to \textit{characterize them}. These cores correspond to the subset of vertices that stays constant, partition-wise, over all solutions constituting a class. Note that our workflow is relatively generic, in the sense that one could apply it to another optimization problem, provided steps 2 and 4 are adjusted to fit the nature of the solutions. In the rest of this section, we review the different steps of our framework in detail.

\subsection{Enumerating All Optimal Solutions}
\label{subsec:EnumOptPartitions}
The enumeration of all distinct optimal solutions can be very time- and memory-consuming, so we need an efficient method.
We handle this step by modeling mathematically the CC problem through Integer Linear Programming (ILP)~\cite{Demaine2006} and by applying the method introduced by Danna \textit{et al.}~\cite{Danna2007}. As discussed in Section~\ref{sec:RelatedWork}, this choice is not only efficient among the existing ones, but also problem-independent. Furthermore, it is straightforward to implement it when one can take advantage of a suitably configured industrial optimization solver such as CPLEX. 
Although this combination of ILP model and optimization solver is sufficient to conduct this task, it can still be time-consuming. 
One way to deal with this issue is to strengthen the underlying ILP model through the \textit{cutting plane} approach, as Ales \textit{et al.}~\cite{Ales2016a} do. We adopt the 2-partition and the 2-chorded cycle valid inequalities proposed by Ales \textit{et al}.
Namely, we use these tight valid inequalities (only) during the root relaxation phase, before proceeding to the construction of the search tree.
Adopting this cut strategy improves the processing time during this first step.
We emphasize that, in this work, the enumeration of all solutions is simply a means rather than an end.

\subsection{Computing the Dissimilarity Values}
\label{subsec:ComputingDistance}
At this stage, we have identified all 
optimal solutions associated to the input graph. 
Let us denote as $p$ the number of solutions found.
We now want to gather similar solutions together. For this purpose, we perform a classic cluster analysis, which in turns requires the computation of a dissimilarity matrix. This matrix is obtained by comparing each pair of solutions. The literature contains a number of similarity or dissimilarity measures to perform such a task, each one possessing specific behavior and characteristics~\cite{Meila2015}. 

We use the \textit{Variation of Information} (VI)~\cite{Meila2003}, which was previously selected in numerous applications similar to our context~\cite{Good2010, Karrer2008, Nguyen2015, Fortunato2016}, because it is a true metric in the space of solutions, and possesses appropriate properties (see~\cite{Meila2015} for details). 

\subsection{Performing the Clustering}
\label{subsec:PerformingClustering}


Next, we apply the $k$-medoids clustering method~\cite{Kaufman2009} to our dissimilarity matrix. It is similar to the well-known $k$-means algorithm, in the sense that it partitions the dataset into $k$ clusters, while minimizing the dissimilarity between the members of each cluster and some center of the cluster. The difference is that in $k$-means, this center is an average value, whereas in $k$-medoids it is one of the actual data points from the dataset. The $k$-medoids method is generally used in place of $k$-means when one cannot perform the required average operation, which is true in our case (we cannot straightforwardly process an average partition, i.e. an average optimal solution).

This method requires us to specify $k$, which we do not know in advance. In this situation, the standard approach is to use all possible values of $k$, from $2$ to $p$ (the number of optimal solutions), and assess the quality of the $p-1$ resulting clusterings through some internal criterion. One of the most widespread such measures is the \textit{Silhouette} which characterizes the clustering in terms of internal cohesion and external separation of the modules~\cite{Rousseeuw1987}. It takes a value between $-1$ and $+1$, where the latter represents the best possible clustering. 

In theory, the $k$ value associated with the highest Silhouette is the best candidate. However, in practice, one possibly has to set a threshold value large enough to ensure a reasonable cluster structure. Obtaining a Silhouette score above this threshold indicates that each cluster contains very similar solutions, and is at the same time clearly separated from the others. Otherwise, a Silhouette score below this threshold means that there is no cluster structure (i.e. a single cluster containing all solutions), or at least that the clustering is inconclusive.
Kaufman \& Rousseeuw recommend to use a threshold value of 0.51 (resp. 0.71) to get a reasonable (resp. strong) cluster structure~\cite{Kaufman2009}. Deciding the value of such a threshold can be considered either as an issue, as it can be a delicate operation, or an advantage, as it allows controlling the strength of the cluster structure. An alternative to determine the existence of a proper cluster structure is to use significance testing.
However, using this type of test requires a certain number of observations, so this approach is not always applicable in practice (as in our case). 

\subsection{Identifying the Core Parts}
\label{subsec:CorePartIdentification}

At the end of the previous step, we obtain a collection of $k$ clusters, each corresponding to a class of solutions that are, by construction relatively similar. We now want to assess how different these classes can be. For this purpose, we leverage the concept that we call \textit{core part}. The core part of a class is the maximal subset of vertices whose relative module assignment stays constant over all the solutions constituting the class. When two vertices belong to the core part, we call them \textit{core vertices}, and they are either always in the same module, or always in different modules, for all the solutions of the class. 
Consequently, vertices that are always \textit{isolated} (i.e. that constitute their own module) are core vertices, as their module assignment always differ from the rest of the core part. 
It is also possible to obtain an \textit{overall core part} by proceeding similarly with all the solutions in the space (i.e. not focusing on a single class).

To identify the core part of a class, we rely on the idea of \textit{consensus matrix} (a.k.a. co-association matrix) originating from Consensus Clustering~\cite{Lancichinetti2012}. The consensus matrix $C$ of a class is an $n \times n$ matrix, whose entry $C_{ij}$ indicates the number of solutions in which vertices $v_i$ and $v_j$ are assigned to the same module, divided by the total number of solutions constituting the solution class. Entries equal to one indicate vertices that are always assigned together to the same modules over all solutions. 


 



\section{Dataset}
\label{sec:Experiments}
This section is dedicated to the description of the dataset used in our experiments. We first define the random model that we propose to generate complete signed graphs (Section~\ref{subsec:network-generation}), before detailing the properties of the generated graphs (Section~\ref{subsec:data}).

\subsection{Random Model}
\label{subsec:network-generation}

To answer our initial questions, we apply our framework to synthetic signed networks generated using a random model.
As mentioned in the introduction, in this article we focus on \textit{complete} \textit{unweighted} graphs, the simplest form of signed graphs, and leave incomplete and/or weighted graphs to future work. Application-wise, complete unweighted signed networks are much less used in the literature, but they nevertheless fit certain modeling situations and methodological choices. For the unweighted aspect, it can be that the studied relations are better represented by binary values (e.g. alliance/conflict between countries in international relationships~\cite{Doreian2015}), or that the authors prefer to use such values for practical reasons (e.g. limited or unreliable information~\cite{Esteban2012}). Regarding the completeness of the network, it can be an artefact of the extraction process~\cite{Ilany2013}, but it can alternatively reflect the nature of the modeled system. This is for example the case of certain networks representing voting behaviors~\cite{Kropivnik1996, Arinik2019}. More generally, authors tend to derive complete signed graphs when they work with similarity (e.g.~\cite{Ahmadian2020}) and correlation (e.g.~\cite{Harary2002}) matrices, in order to use CC to perform a form of cluster analysis.

As showed by our examples from Section~\ref{subsec:BackgroundMultiplicity}, their simplicity does not prevent complete unweighted signed graphs to exhibit cases of multiple and structurally very different solutions. Moreover, the literature shows that this can also be the case in sparse graphs~\cite{Davis1967, Doreian1996, Doreian2005}, so this simplicity is not the cause of this property either. Their simplicity makes complete unweighted graphs particularly suitable in the context of this article, which is the first step of longer-term work. Indeed, it allows us to focus on the most \textit{essential} parameters when randomly generating graphs to constitute our dataset, and later when studying their effect on the solution space of CC. Put differently, it allows us to postpone the study of issues related to sparse graphs, such as density, degree distribution, and proportion of negative edges. That is not to say that these parameters do not affect the solution space of CC, but rather that they require their own study, in complement to the present one.

We propose a simple yet principled random model designed to produce complete unweighted networks with built-in modular structure. Its R implementation is publicly available online\footnote{\url{https://github.com/CompNet/SignedBenchmark}}. This model relies on only three parameters: $n$ (number of vertices), $\ell_0$ (initial number of modules) and $q_{m}$ (proportion of misplaced edges, i.e. edges meant to be frustrated by construction).
First, we produce a graph containing $n$ vertices, divided into $\ell_0$ approximately equal-sized modules to form a partition $P_0$. 
We connect them with negative and positive edges, in such a way that this complete graph is perfectly balanced. Second, we introduce some imbalance into the graph, so as to match parameter $q_{m}$. For this purpose, we randomly select a pair of well-placed negative and positive edges, then switch their signs in order to make both of them misplaced. The process is repeated to other pairs of edges. On the one hand, this mechanism causes a restriction on the upper bound of $q_{m}$. But on the other hand, it allows preserving the ratio of positive 
to negative edges 
in the graph, and therefore avoids introducing another parameter in the model. It is important to note that the detected graph imbalance $I(P)$ does not necessarily match $q_{m}$, as it may be possible to find a better partition $P$ than the initial $P_0$ due to the introduction of imbalance.



\subsection{Generated Data}
\label{subsec:data}
It is well known that exact approaches solving most clustering problems (including ours) do not scale well, even when looking for a \textit{single} optimal solution, due to their NP-hard nature. In that respect, according to our preliminary tests, our strengthened ILP model can handle CC on complete graphs containing up to approximately 80 vertices in a reasonable time. However, in our case we must look for \textit{all} optimal solutions in order to retrieve the complete solution space of each considered instance of the problem. Enumerating all optimal solutions through CPLEX requires a large amount of RAM and a long execution time, which lowers the maximum network order that we can practically handle to $36$ vertices. It is worth noting that this graph order is on par with the other works dealing with spaces of optimal solutions (e.g. up to 20 geographic units in \cite{Arthur1997}).
Moreover, as far as our experiments go, our results from Section~\ref{subsec:ResultsNbSol} show no noticeable effect of the graph order on the number of solutions --whether this holds for larger graphs remains to be tested, though.

Our experiments are conducted on random instances with a built-in modular structure, where $\ell_0 \in \{2, 3, 4\}$ and $n \in \{16, 20, 24, 28, 32, 36\}$. For replication, this generation process is repeated $100$ times for each parameter set. In total, we produce $10,200$ instances for $\ell_0=2$; $7,000$ instances for $\ell_0=3$; and $5,000$ instances for $\ell_0=4$, which makes a total of $22,200$ instances. All these data as well as the solutions we identified are publicly available  online\footnote{\url{https://doi.org/10.6084/m9.figshare.8233340}\label{foot:figshare}}. 


\section{Results}
\label{sec:Results}
We now investigate the space of optimal solutions. We first consider the generated dataset (Sections~\ref{subsec:ResultsNbSol} to~\ref{subsec:ResultsCoreParts}): we present the results in the order that our workflow follows (see Section~\ref{sec:Methods}), since it is a pipeline. In addition, we apply our method to a small network of international relations (Section~\ref{subsec:ResultsSyria}) to show its relevance on real-world data.

Regarding the synthetic graphs, we present a selection of the most relevant results in Figures~\ref{fig:nb-sol-qm-k2-k3-k4} 
to \ref{fig:class-core-part-size-k2-k3-k4},
for $\ell_0 \in \{2, 3, 4\}$.
The complete results\footref{foot:figshare} as well as our source code\footnote{\url{https://github.com/CompNet/Sosocc}} are available online, though. We first describe these plots generically here, for matters of convenience, before interpreting them. In these figures there are 3 subfigures (identified by a letter: a, b, c). Each subfigure is a block of 6 plots, and focuses on a specific variable of interest, represented on the $y$-axis of the plots. The $x$-axis can either represent parameter $q_{m}$ (Figures~\ref{fig:nb-sol-qm-k2-k3-k4}, \ref{fig:imb-perc-k2-k3-k4}, \ref{fig:nb-module-qm-k2-k3-k4}), or the \textit{detected} graph imbalance $I(P)$ 
(Figures~\ref{fig:nb-sol-realImb-k2-k3-k4}, \ref{fig:single-class-prop-k2-k3-k4}, \ref{fig:class-core-part-size-k2-k3-k4}). 
Each plot in a subfigure corresponds to a different graph order $n$ (number of vertices). The plots in Figure~\ref{fig:single-class-prop-k2-k3-k4} represent the data as histograms, whereas the others contain violin plots, each one representing the results from $100$ replications for the same parameter set. 
In each violin plot, the interquartile range is 
shown as a purple thick line, the mean as a green triangle and the median as a blue dot. 
In case of a unique value, only the mean and median appear.




\subsection{Number of Solutions}
\label{subsec:ResultsNbSol}

\begin{table}[h]
	\caption{Average and maximal numbers of optimal solutions obtained over $100$ replications, for $\ell_0=2$. 
	}
    \small
	\centering
    \setlength{\tabcolsep}{0.25em} 
    \begin{tabular}{|r | r r | r r | r r | r r | r r | r r |}
    	\hline
    	\multirow{2}{*}{\diagbox{$q_{m}$}{$n$}} &
        \multicolumn{2}{c|}{\textbf{16}} & \multicolumn{2}{c|}{\textbf{20}} & \multicolumn{2}{c|}{\textbf{24}} & \multicolumn{2}{c|}{\textbf{28}} & \multicolumn{2}{c|}{\textbf{32}} & \multicolumn{2}{c|}{\textbf{36}} \\
            & Average & Max. & Average & Max. & Average & Max. & Average & Max. & Average & Max. & Average & Max. \\
        \hline
        \textbf{0.05} & 1.00 & 1 & 1.00 & 1 & 1.00 & 1 & 1.00 & 1 & 1.00 & 1 & 1.00 & 1 \\
        \textbf{0.10} & 1.00 & 1 & 1.00 & 1 & 1.00 & 1 & 1.00 & 1 & 1.00 & 1 & 1.00 & 1 \\
        \textbf{0.15} & 1.09 & 3 & 1.01 & 1 & 1.00 & 1 & 1.00 & 1 & 1.00 & 1 & 1.00 & 1 \\
        \textbf{0.20} & 1.33 & 5 & 1.13 & 3 & 1.10 & 4 & 1.02 & 2 & 1.03 & 2 & 1.00 & 1 \\
        \textbf{0.25} & 2.71 & 23 & 1.95 & 6 & 1.71 & 18 & 1.35 & 4 & 1.32 & 6 & 1.15 & 6 \\
        \textbf{0.30} & 4.63 & 36 & 4.04 & 20 & 3.23 & 17 & 3.49 & 19 & 2.51 & 16 & 2.25 & 11 \\
        \textbf{0.35} & 7.02 & 119 & 6.91 & 113 & 5.75 & 75 & 5.12 & 33 & 5.73 & 39 & 5.93 & 72 \\
        \textbf{0.40} & 6.16 & 69 & 6.94 & 54 & 5.79 & 41 & 6.98 & 39 & 8.75 & 64 & 6.80 & 60 \\
        \textbf{0.45} & 5.77 & 34 & 5.83 & 38 & 9.58 & 151 & 6.31 & 44 & 6.94 & 72 & 5.33 & 34 \\
        \textbf{0.50} & 5.87 & 45 & 6.89 & 46 & 5.38 & 51 & 9.39 & 135 & 5.63 & 43 & 9.82 & 159 \\
        \textbf{0.55} & 6.32 & 150 & 5.90 & 73 & 5.85 & 28 & 6.99 & 58 & 7.02 & 47 & 6.27 & 64 \\
        \textbf{0.60} & 6.05 & 53 & 6.23 & 71 & 7.66 & 45 & 8.72 & 91 & 5.98 & 36 & 9.41 & 116 \\
        \textbf{0.65} & 8.23 & 74 & 5.99 & 56 & 8.59 & 93 & 5.87 & 52 & 7.31 & 62 & 8.28 & 49 \\
        \textbf{0.70} & 6.79 & 48 & 9.18 & 75 & 6.67 & 107 & 6.74 & 50 & 11.42 & 182 & 5.94 & 31 \\
        \textbf{0.75} & 8.62 & 78 & 6.95 & 84 & 11.42 & 171 & 12.3 & 204 & 9.30 & 108 & 10.52 & 68 \\
        \textbf{0.80} & 20.01 & 361 & 21.53 & 720 & 17.05 & 202 & 17.1 & 183 &  17.93 & 321 & 14.82 & 127 \\
        \textbf{0.85} & 77.07 & 2,948 & 63.37 & 1,488 & 43.13 & 917 & 37.21 & 473 & 30.18 & 435 & 31.00 & 347 \\
        \textbf{0.90} & 4,946.40 & 10,009 & 3,610.40 & 13,403 & 10,811.50 & 71,875 & 2,529.40 & 8,150 & 7,196.00 & 65,667 & 588.30 & 2,767 \\
        \hline
	\end{tabular}
	\label{tab:average-max-nb-sol-k2}
\end{table}        

\begin{figure}[!htb]
\captionsetup{width=0.9\textwidth}
    \subfloat{
        \centerline{\includegraphics[height=0.200\textheight,clip=true]{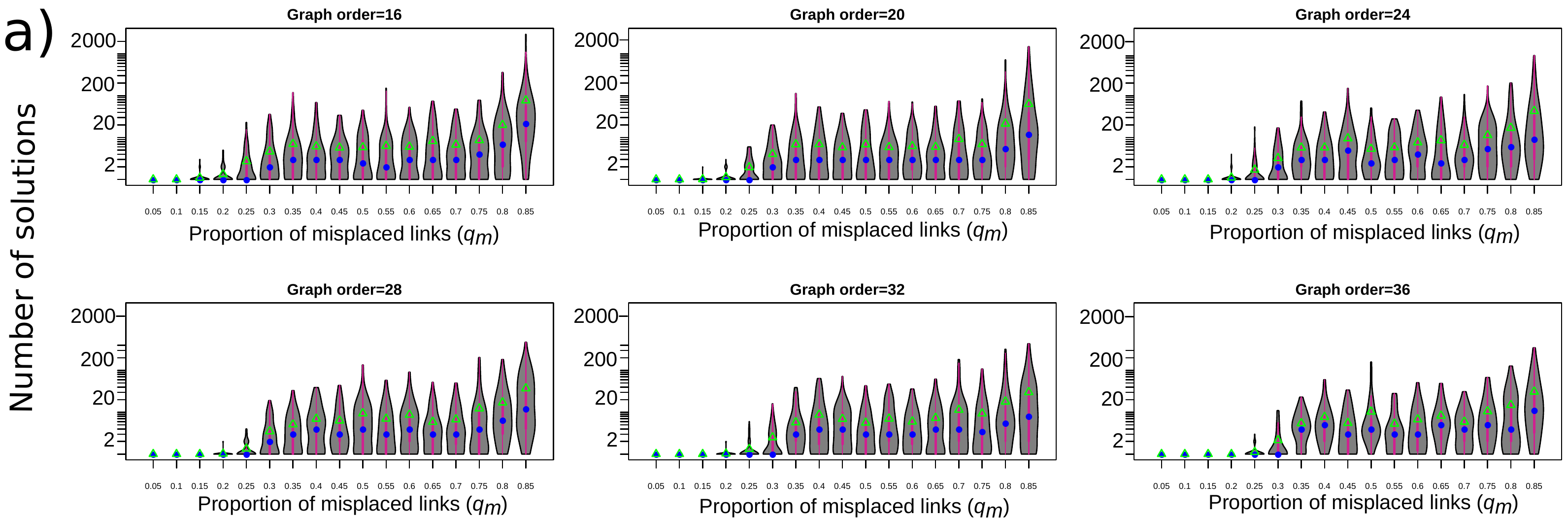}}
        \label{fig:nb-sol-qm-k0=2}
    }\\
    \centerline{\subfloat{
        \includegraphics[height=0.275\textheight,clip=true]{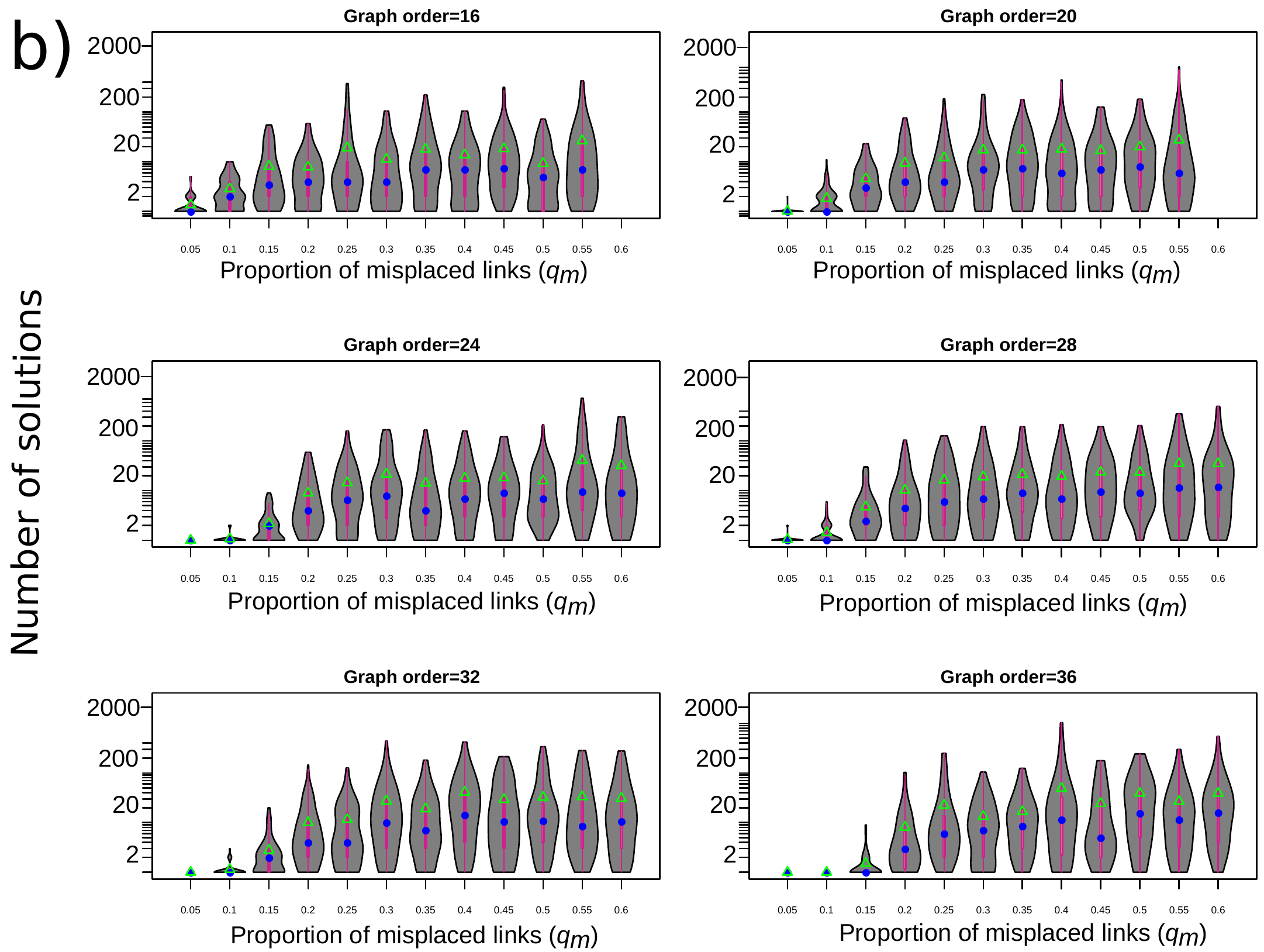}
        \label{fig:nb-sol-qm-k0=3}
    }
    \subfloat{
        \includegraphics[height=0.275\textheight,clip=true]{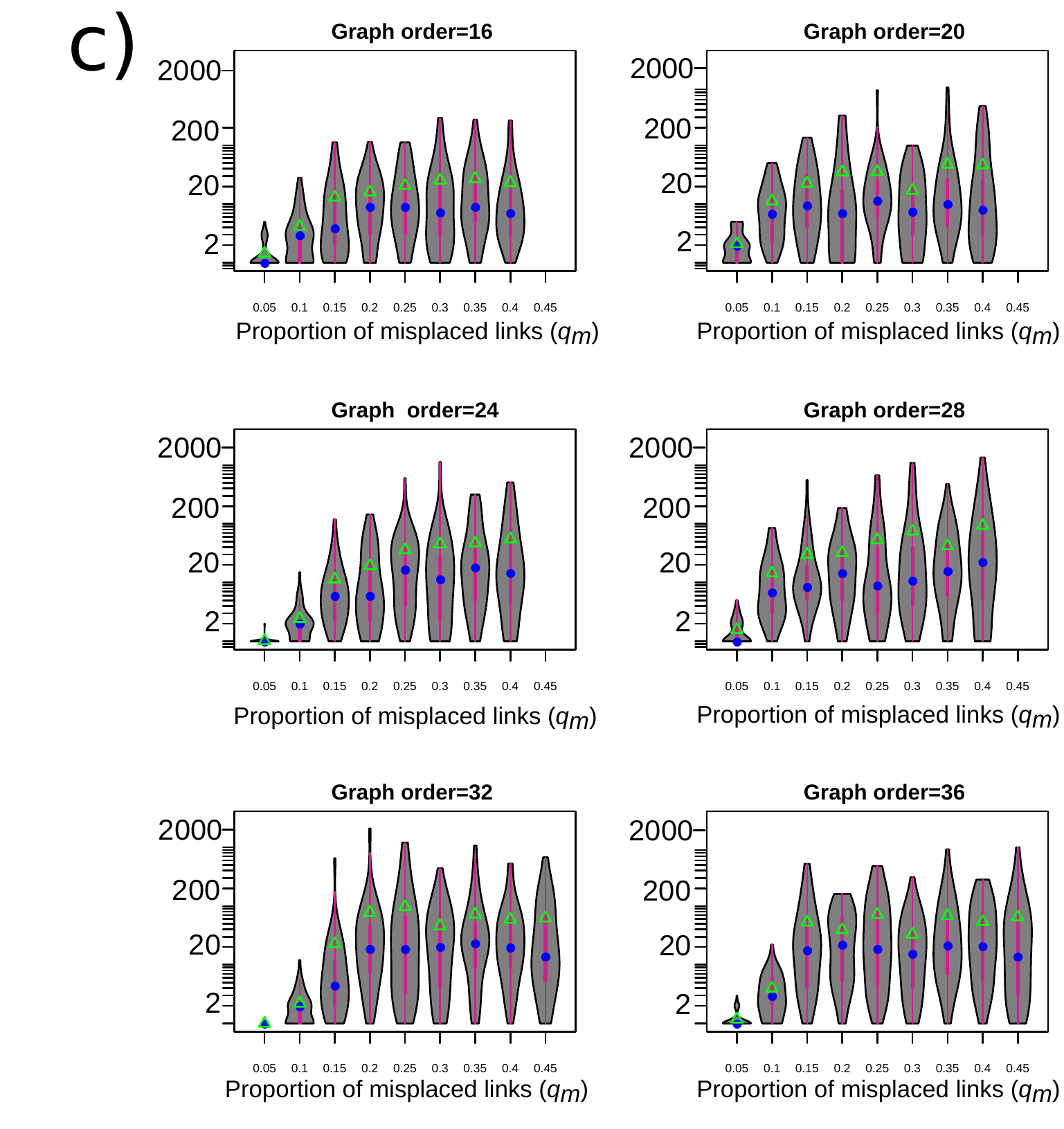}
        \label{fig:nb-sol-qm-k0=4}
    }}
    \caption{Number of solutions (log-scaled) as a function of $q_m$, (a) for $\ell_0=2$, (b) for $\ell_0=3$ and (c) for $\ell_0=4$. Notice that an $x$-axis value may be empty if the parameter set is not defined or no data is available. Plots available at \href{https://doi.org/10.6084/m9.figshare.8233340}{10.6084/m9.figshare.8233340} under CC-BY license.}
    \label{fig:nb-sol-qm-k2-k3-k4}
\end{figure}

We first study how frequent multiple optimal solutions are. Subfigures~\ref{fig:nb-sol-qm-k0=2}, \ref{fig:nb-sol-qm-k0=3} and \ref{fig:nb-sol-qm-k0=4} show the number of optimal solutions as a function of $q_{m}$, for different graph orders $n$, and for $\ell_0=2$, $3$ and  $4$, respectively. Note that the $y$-axis uses a logarithmic scale to cope with the fast growth of the number of solutions. We observe that for all $\ell_0$ values and a small $q_{m}$, there is a unique solution in most of the cases. Nonetheless, when $q_{m}$ increases, i.e. when we introduce more misplaced edges, multiple optimal solutions are more and more frequent (see Table~\ref{tab:average-max-nb-sol-k2}). There is a unique optimal solution in $45\%$ of the graph instances generated for $\ell_0=2$, $28\%$ for $\ell_0=3$ and $21\%$ for $\ell_0=4$. These proportions, completed by visual inspection, indicate that there are more optimal solutions when $\ell_0$ increases, despite the upper bound restriction of $q_{m}$ mentioned in Section~\ref{subsec:network-generation}.

\begin{figure}[!htb]
\captionsetup{width=0.9\textwidth}
    \centerline{\subfloat{
        \includegraphics[height=0.275\textheight]{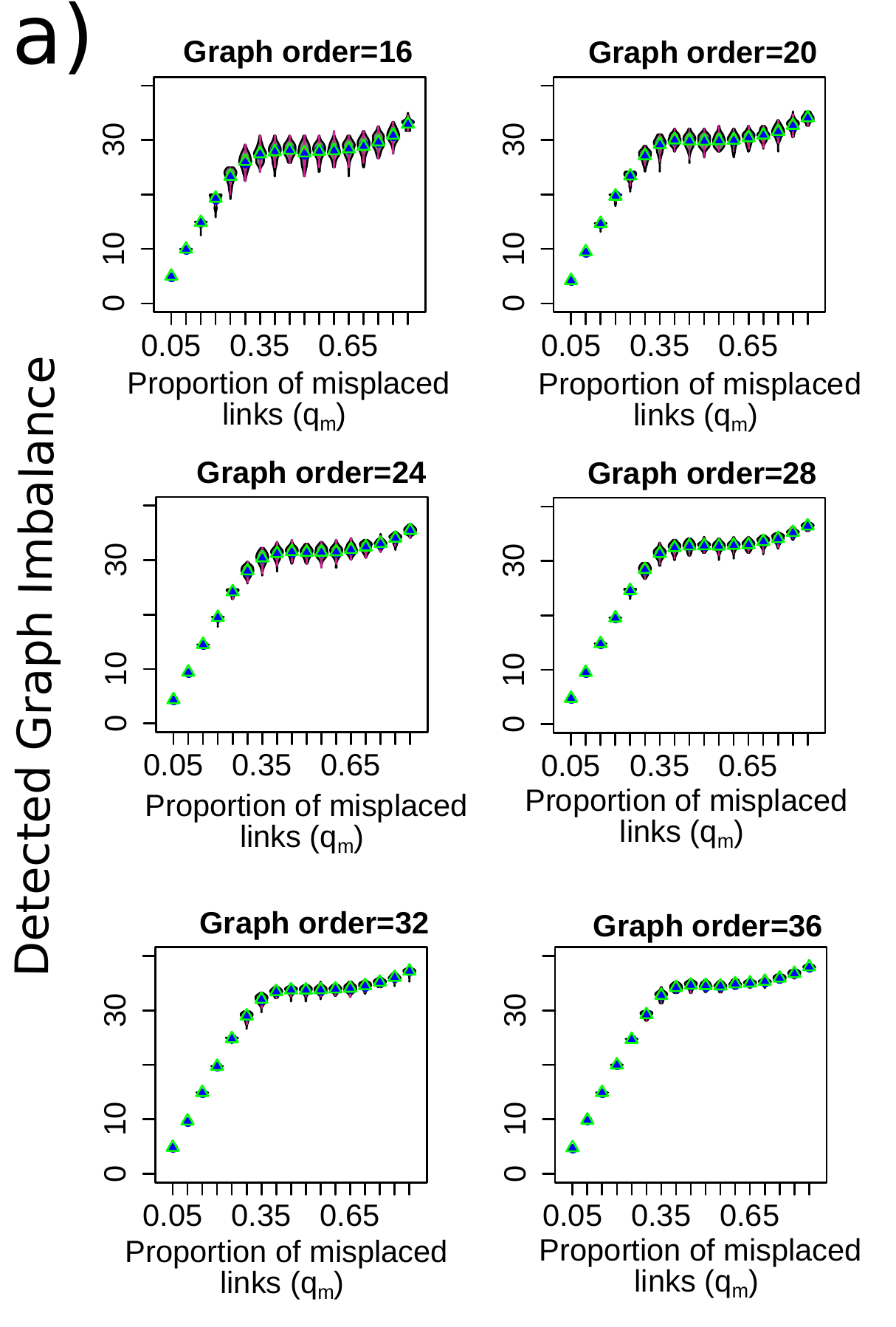}
        \label{fig:imb-perc-k0=2}
    }
    \subfloat{
        \includegraphics[height=0.275\textheight]{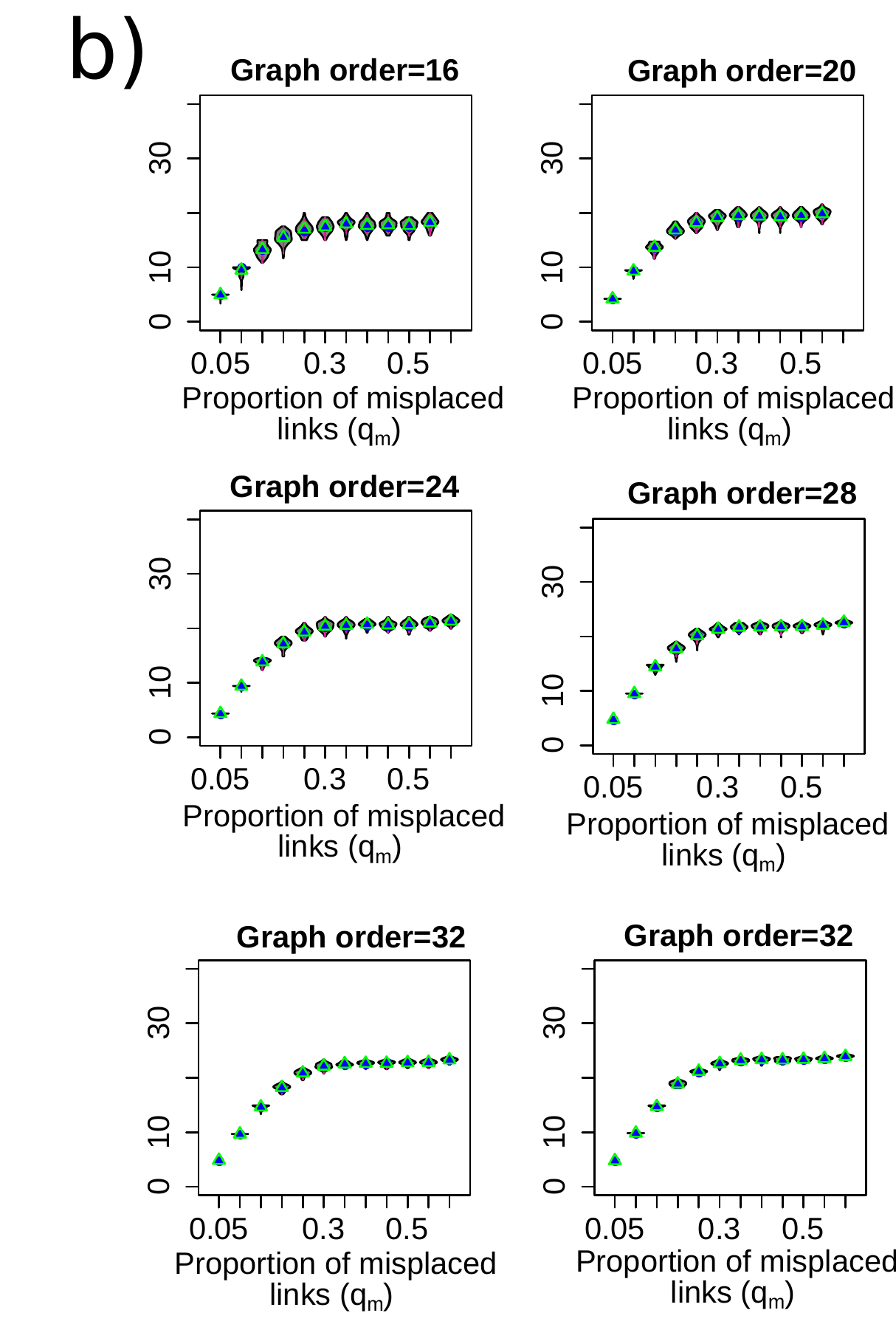}
        \label{fig:imb-perc-k0=3}
    }
    \subfloat{
        \includegraphics[height=0.275\textheight]{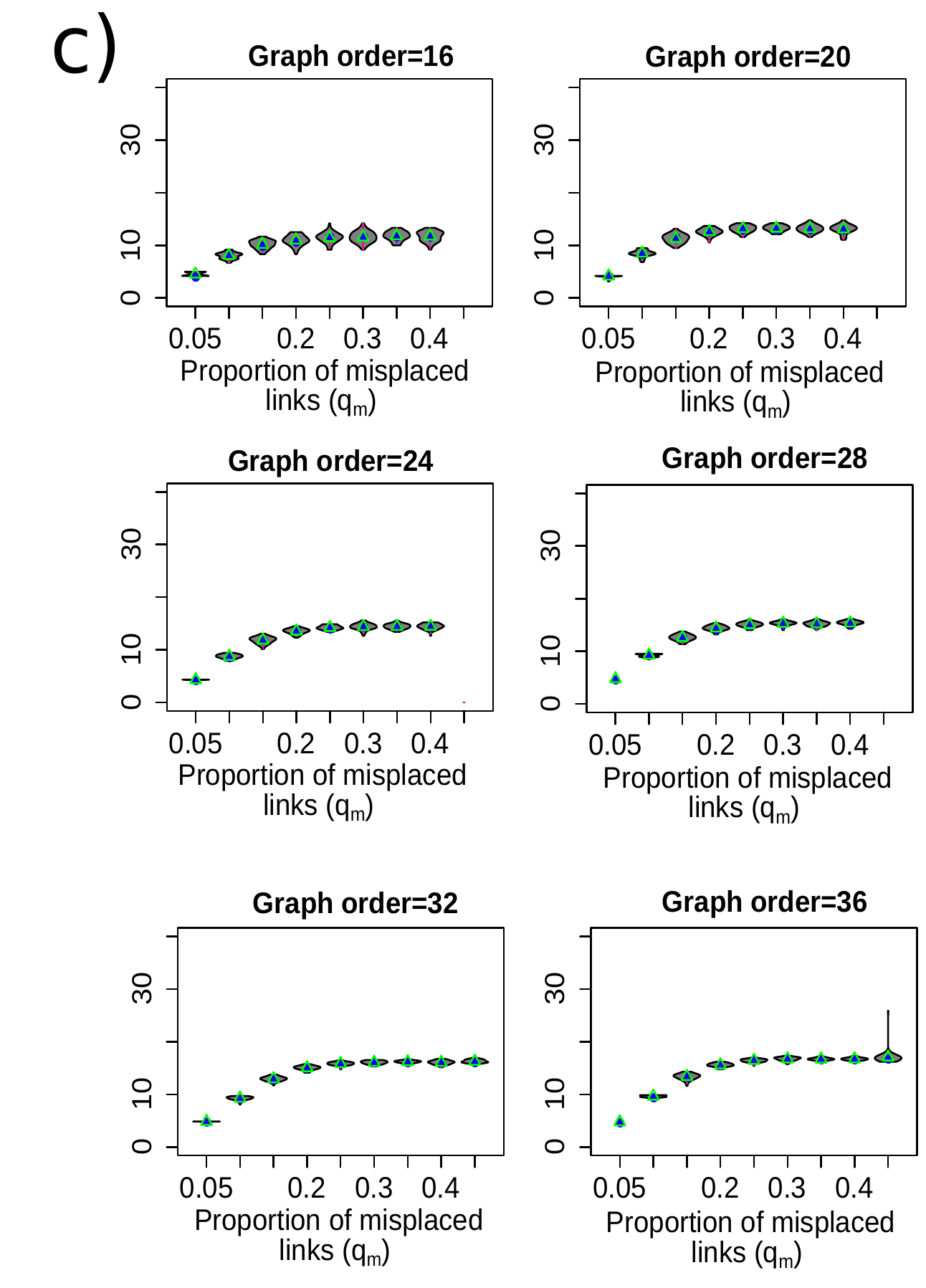}
        \label{fig:imb-perc-k0=4}
    }}
    \caption{Detected graph imbalance $I(P)$ as a function of $q_m$, (a) for $\ell_0=2$, (b) for $\ell_0=3$ and (c) for $\ell_0=4$. Notice that an $x$-axis value may be empty if the parameter set is not defined or no data is available. Plots available at \href{https://doi.org/10.6084/m9.figshare.8233340}{10.6084/m9.figshare.8233340} under CC-BY license.}
    \label{fig:imb-perc-k2-k3-k4}
\end{figure}

\begin{figure}[!htb]
    \subfloat{
        \centerline{\includegraphics[height=0.200\textheight,clip=true]{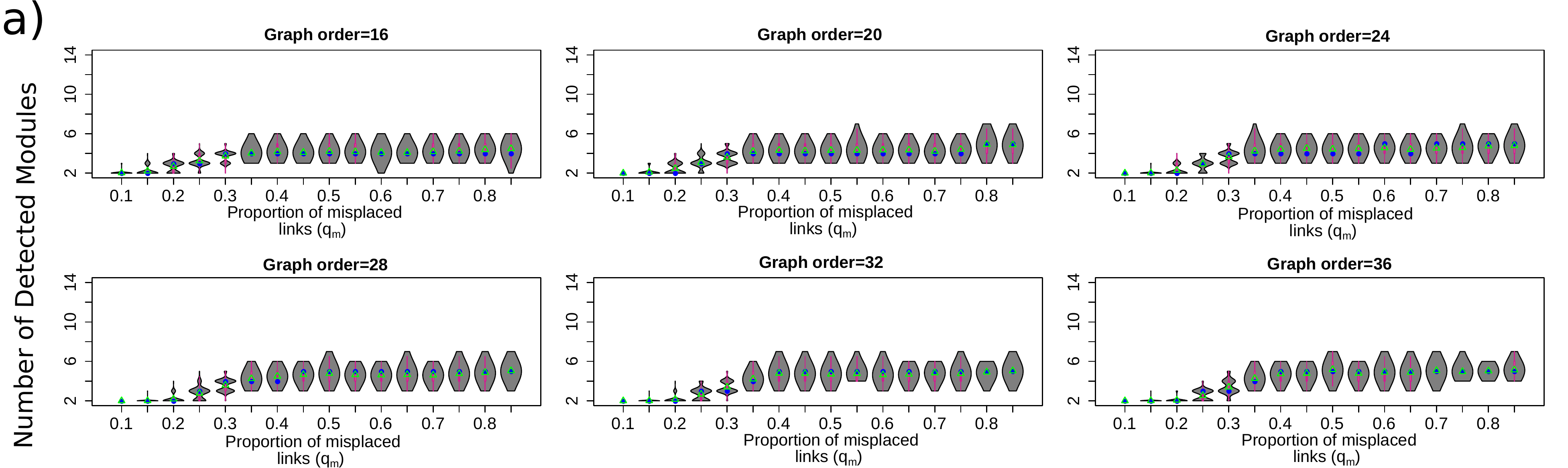}}
        \label{fig:nb-module-qm-k0=2}
    }\\
    \centerline{\subfloat{
        \includegraphics[height=0.325\textheight,clip=true]{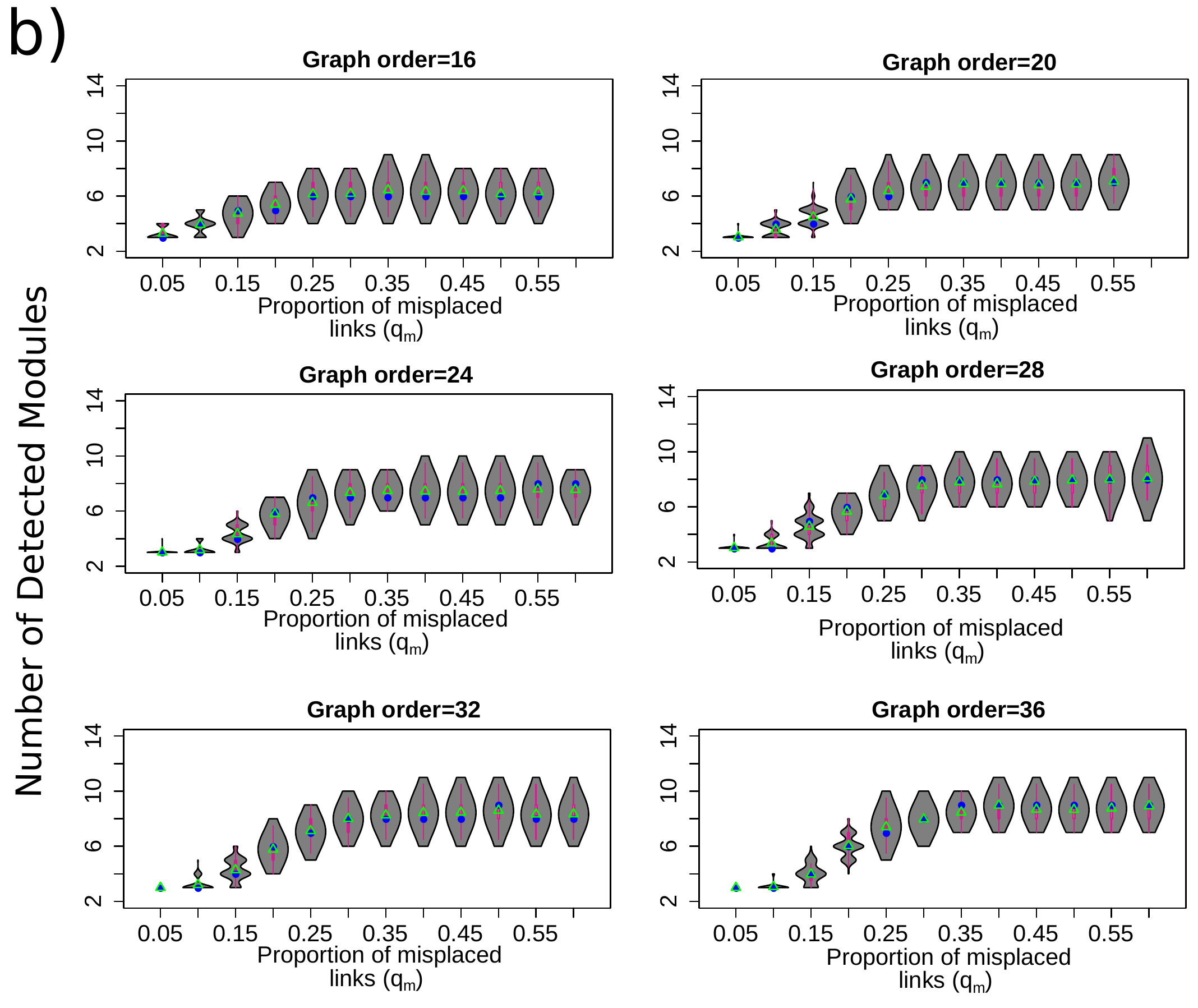}
        \label{fig:nb-module-qm-k0=3}
    }
    \subfloat{
        \includegraphics[height=0.325\textheight,clip=true]{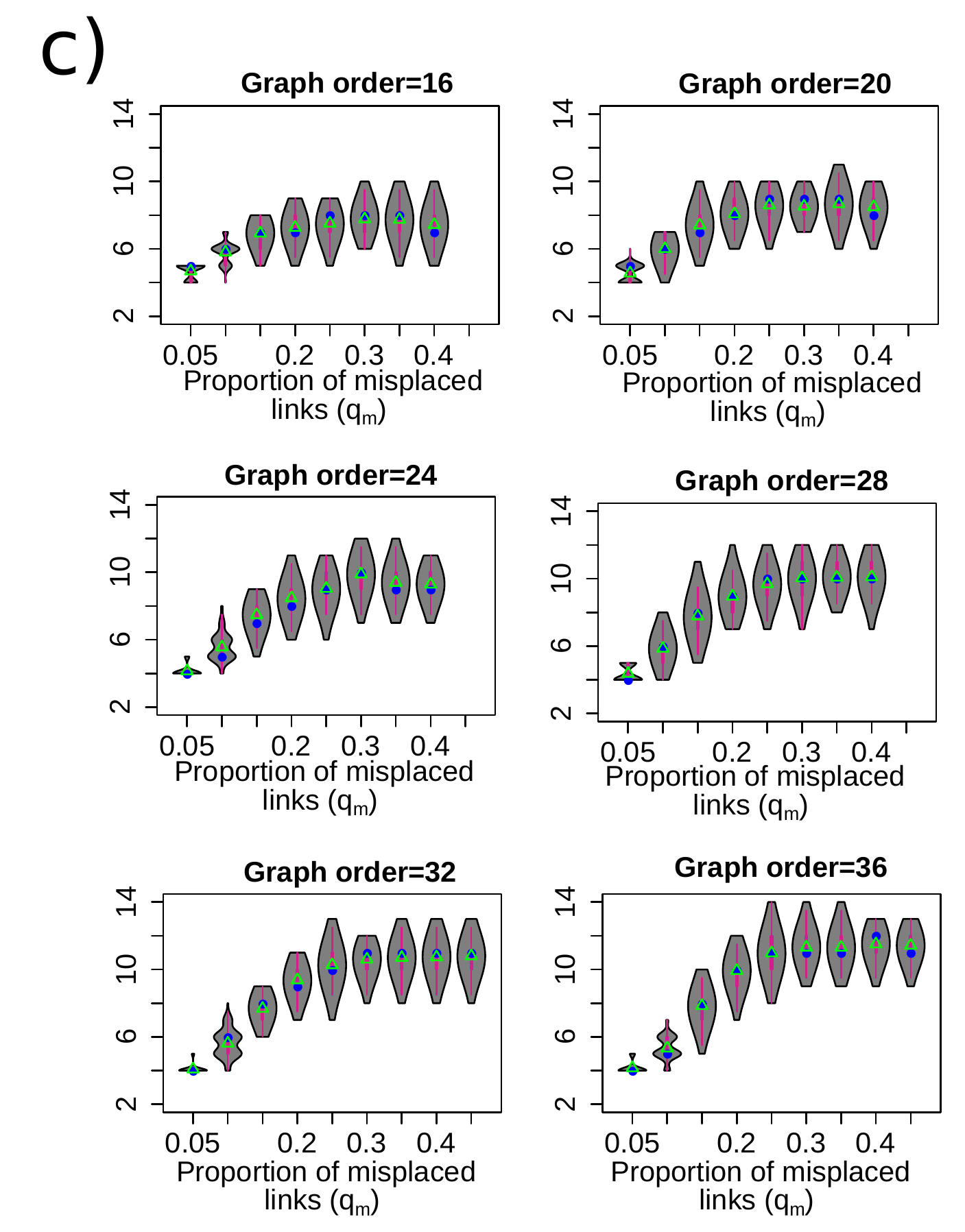}
        \label{fig:nb-module-qm-k0=4}
    }}
    \caption{Number of detected modules as a function of $q_m$, (a) for $\ell_0=2$, (b) for $\ell_0=3$, and (c) for $\ell_0=4$. Notice that an $x$-axis value may be empty if the parameter set is not defined or no data is available. Plots available at \href{https://doi.org/10.6084/m9.figshare.8233340}{10.6084/m9.figshare.8233340} under CC-BY license.}
    \label{fig:nb-module-qm-k2-k3-k4}
\end{figure}

This fact can be explained by considering the detected graph imbalance $I(P)$ as a function of $q_m$, as illustrated in Figure~\ref{fig:imb-perc-k2-k3-k4}.
For all $\ell_0$ values, we observe that when $q_{m}$ increases, $I(P)$ also increases for small $q_{m}$ values, but then reaches a plateau. Yet, one would expect the imbalance to directly depend on the number of misplaced edges introduced in the graph. However, when $q_{m}$ exceeds some threshold, the number of misplaced edges (relative the initial partition) becomes so large that it provides some form of flexibility to graph partitioning. Consequently, even if these misplaced edges are randomly distributed, it becomes possible to partition the graph into a larger number of smaller modules allowing to reach a lower imbalance than expected (though still high). In addition, this flexibility also allows finding several equally good partitions into such small modules, which leads to multiple optimal solutions. This is illustrated in Figure~\ref{fig:nb-module-qm-k2-k3-k4}, which displays the number of detected modules as a function of the number of misplaced edges $q_m$. One can observe an increase in the number of detected modules and/or in their dispersion when $q_m$ increases, up to a certain point. We also note that this effect is stronger when the initial number of modules $\ell_0$ increases.

\begin{figure}[!htb]
\captionsetup{width=0.9\textwidth}
    \subfloat{
        \centerline{\includegraphics[height=0.200\textheight,clip=true]{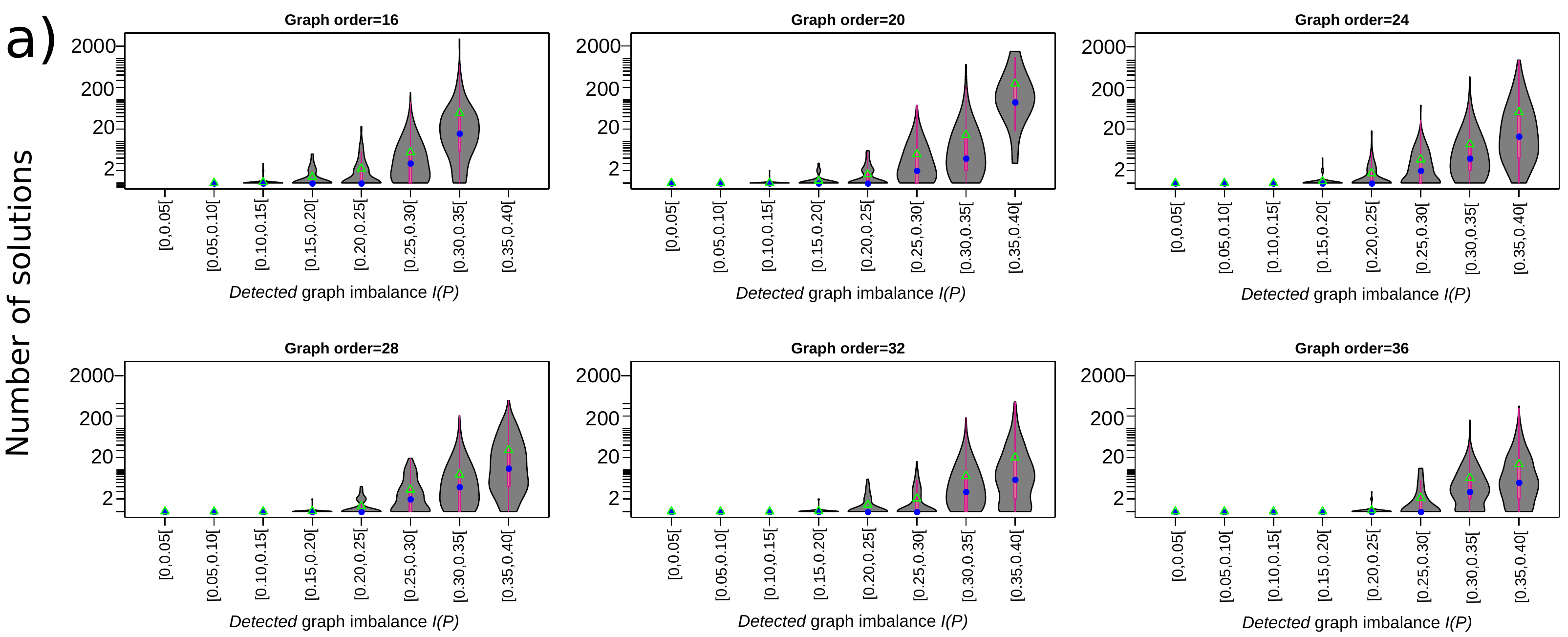}}
        \label{fig:nb-sol-realImb-k0=2}
    }\\
    \centerline{\subfloat{
        \includegraphics[height=0.275\textheight,clip=true]{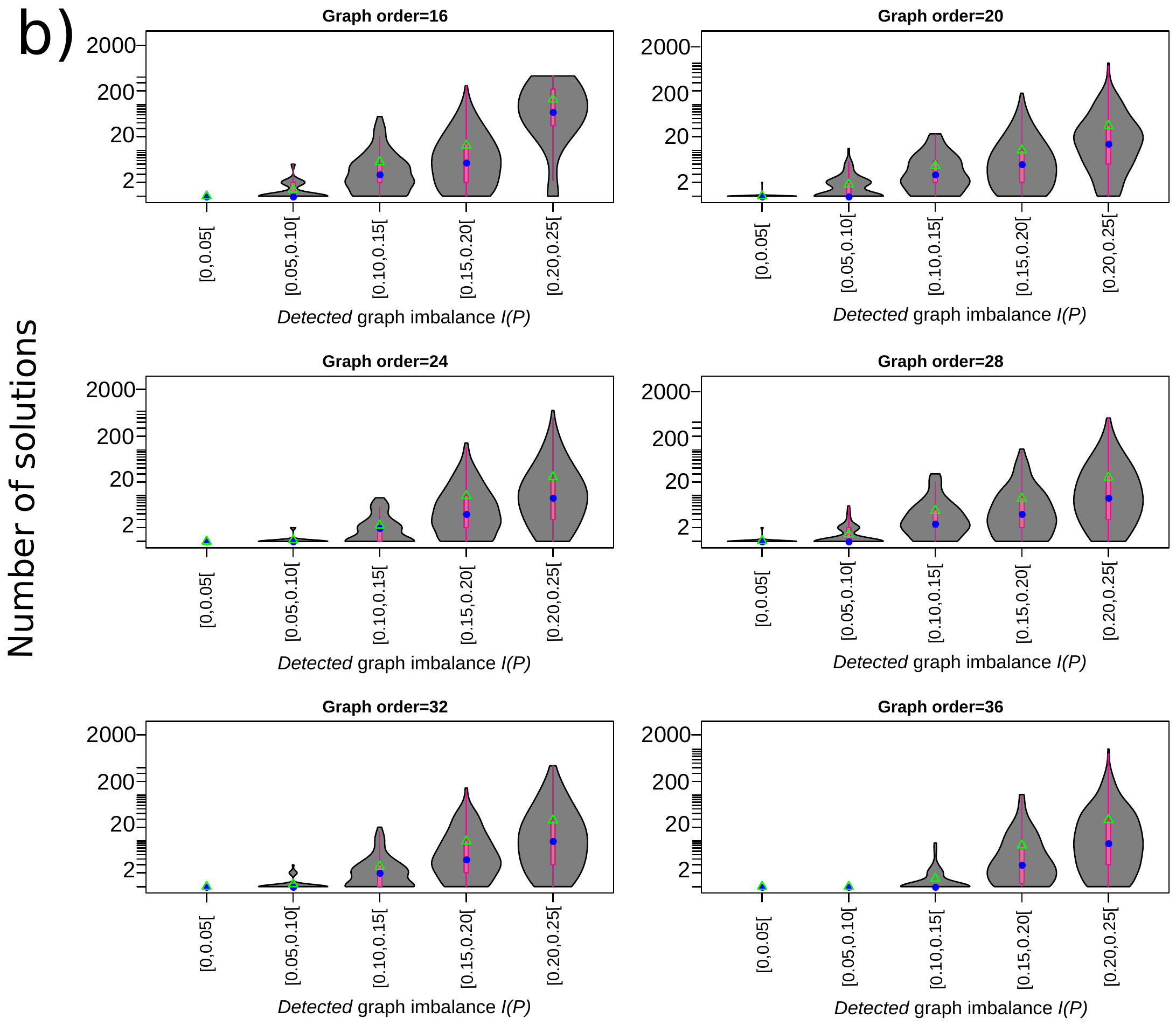}
        \label{fig:nb-sol-realImb-k0=3}
    }
    \subfloat{
        \includegraphics[height=0.275\textheight,clip=true]{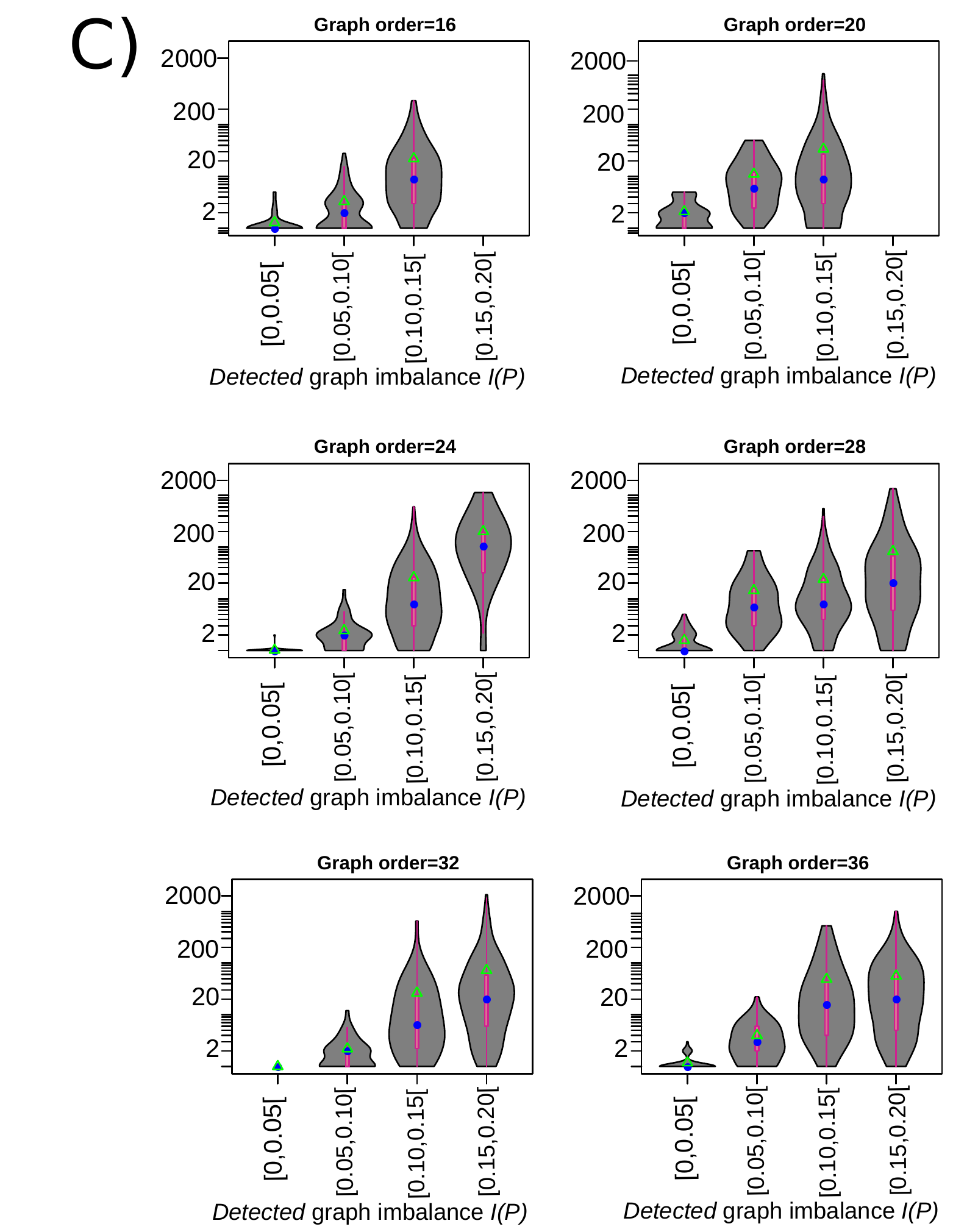}
        \label{fig:nb-sol-realImb-k0=4}
    }}
    \caption{Number of solutions (log-scaled) as a function of $I(P)$, (a) for $\ell_0=2$, (b) for $\ell_0=3$ and (c) for $\ell_0=4$. Notice that an $x$-axis value may be empty if the parameter set is not defined or no data is available. Figures available at \href{https://doi.org/10.6084/m9.figshare.8233340}{10.6084/m9.figshare.8233340} under CC-BY license.}
    \label{fig:nb-sol-realImb-k2-k3-k4}
\end{figure}

We expected the order of the graph to affect the number of solutions, as one could suppose that a larger graph offers more possible partitions. However, this does not seem to be the case in our results, at least for $\ell_0=2$, as the trends observed in Subfigure~\ref{fig:nb-sol-qm-k0=2} and Table~\ref{tab:average-max-nb-sol-k2} are very similar for all considered graph orders. There seems to be a slight increase for $\ell_0=3$ and $\ell_0=4$, though, as shown by Subfigures~\ref{fig:nb-sol-qm-k0=3} and \ref{fig:nb-sol-qm-k0=4}, respectively. This is apparent for intermediate values of $q_{m}$, but at this point it is not clear whether this holds for its other values. We adopt a different angle by considering the number of solutions as a function of the \textit{detected} imbalance $I(P)$ in Figure~\ref{fig:nb-sol-realImb-k2-k3-k4}. As for Figure~\ref{fig:nb-sol-qm-k2-k3-k4}, note that the $y$-axis uses a logarithmic scale. This figure confirms that the number of solutions tends to increase with the imbalance, whereas the graph order does not have a clear effect. For instance, when considering $I(P)=[0.20,0.25[$ in Subfigure~\ref{fig:nb-sol-realImb-k0=3}, we see an alternation of increase and decrease in both the average number of solutions and their dispersion when $n$ increases.

To conclude this part, our experiment reveals that it is possible to obtain many optimal solutions when solving the CC problem on certain networks. If the order of the graph does not seem to affect the number of solutions much, on the contrary the graph imbalance certainly plays a key role. A larger imbalance generally leads to more optimal solutions, and in addition our plots show that the dispersion of this number also increases, resulting in extreme values. 
Thus, it certainly seems necessary to assess the multiplicity of solutions in case of relatively imbalanced networks. From a practical point of view though, certain types of real-world networks are known to have a low imbalance~\cite{Leskovec2010}. In this case, identifying all optimal solutions might not seem necessary. But there is no absolute guarantee to get a unique, or even few optimal solutions when the imbalance is low. For instance, we get a maximum of $55$ (and an average of $4.25$) solutions for $\ell_0=3$, $n=16$, $I(P)=[0.10,0.15[$, which is already quite a large number of solutions for such a low imbalance. 
This motivates us to go on with our study and consider the diversity of optimal solutions.

\subsection{Diversity of the Solutions}
\label{subsec:ResultsDissSols}

\begin{figure}[!htb]
\captionsetup{width=0.9\textwidth}
    \centerline{\subfloat{
        \includegraphics[height=0.225\textheight,clip=true]{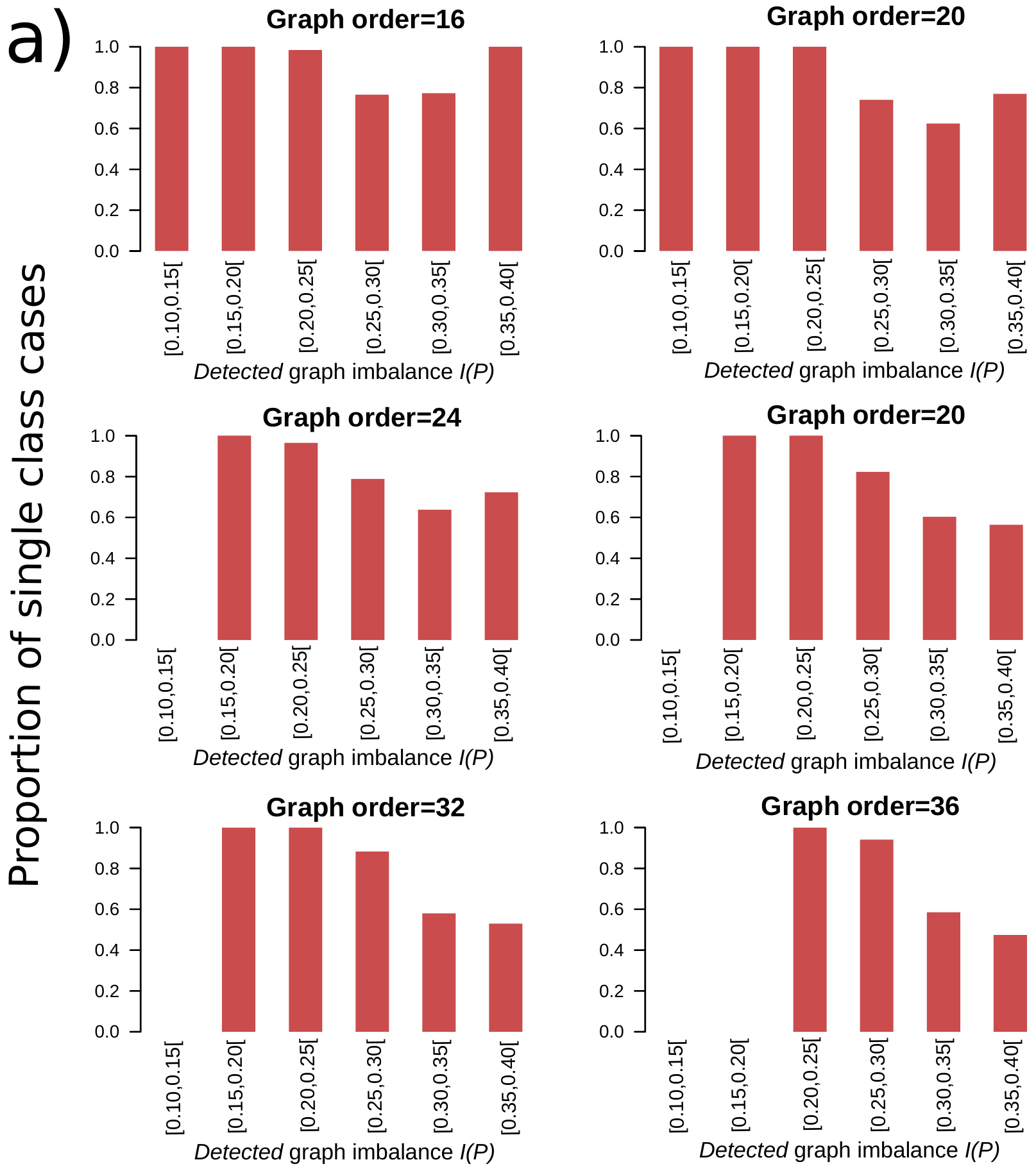}
        \label{fig:single-class-prop-k0=2}
    }
    \subfloat{
        \includegraphics[height=0.225\textheight,clip=true]{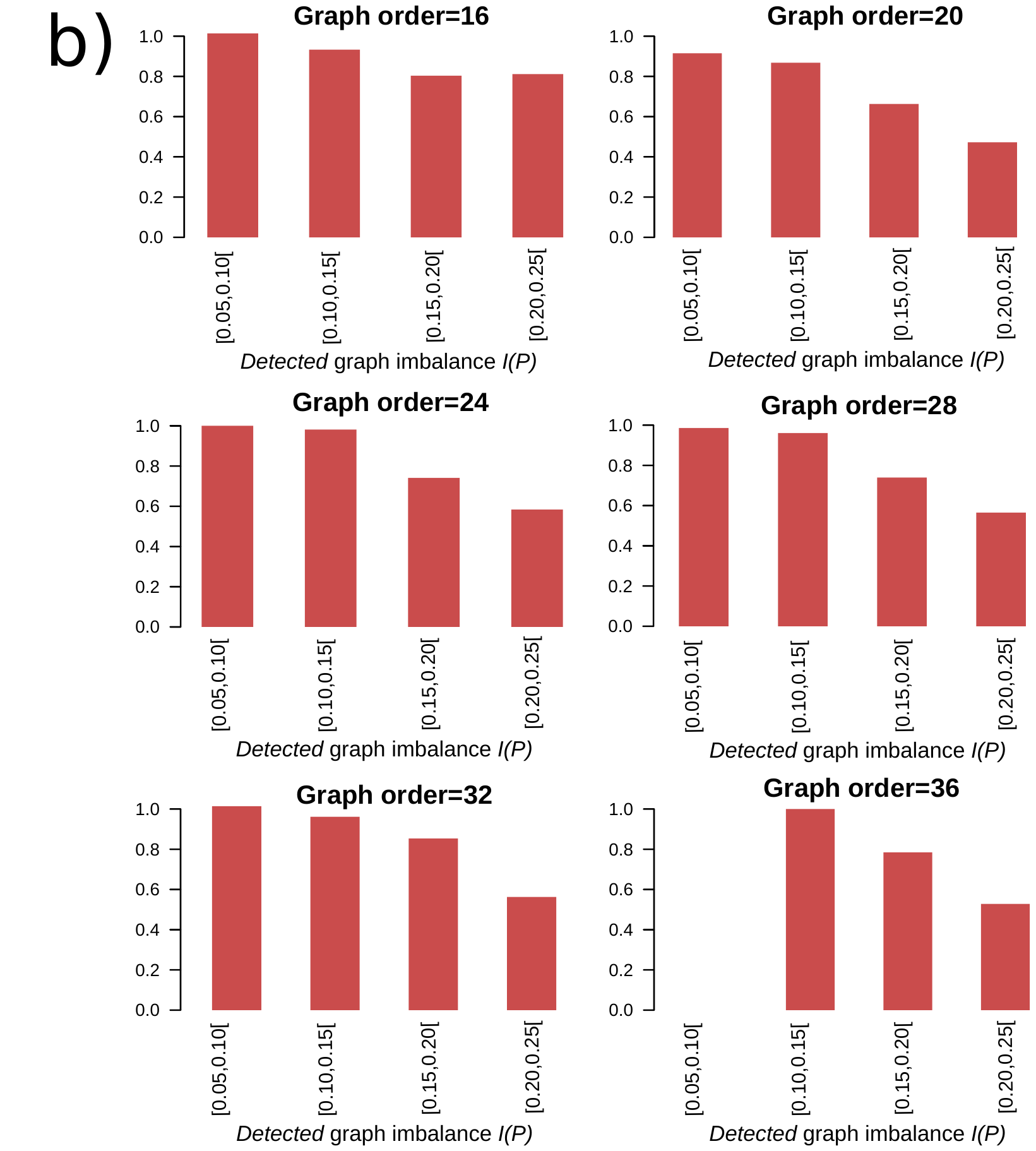}
        \label{fig:single-class-prop-k0=3}
    }
    \subfloat{
        \includegraphics[height=0.225\textheight,clip=true]{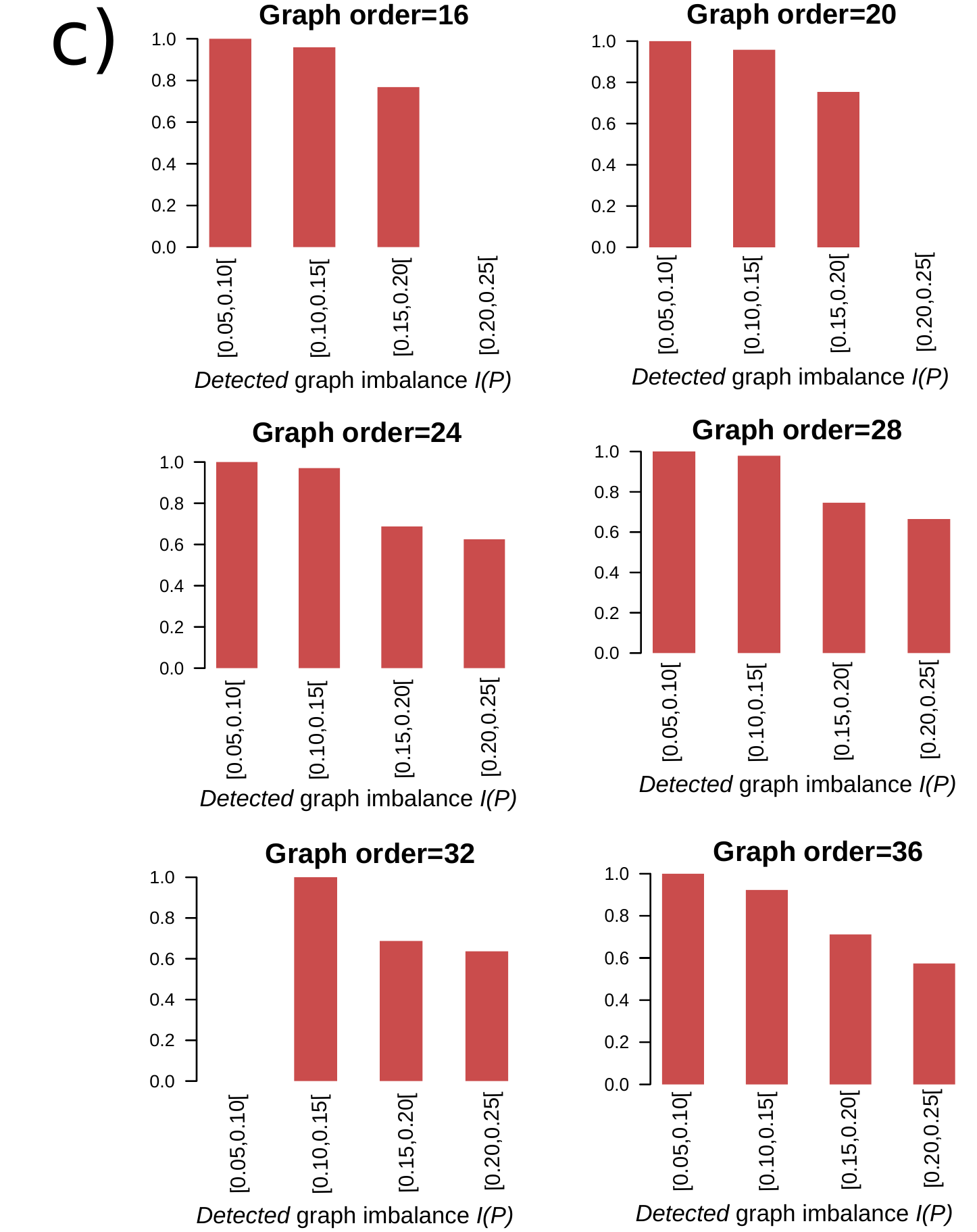}
        \label{fig:single-class-prop-k0=4}
    }}
    \caption{Proportion of single-class cases as a function of the detected imbalance, (a) for $\ell_0=2$, (b) for $\ell_0=3$ and (c) for $\ell_0=4$. Notice that an $x$-axis value may be empty if the parameter set is not defined or no data is available. Plots available at \href{https://doi.org/10.6084/m9.figshare.8233340}{10.6084/m9.figshare.8233340} under CC-BY license.}
    \label{fig:single-class-prop-k2-k3-k4}
\end{figure}

Our second question is how different the obtained solutions are, in case of multiplicity. We answer it by analyzing the numbers of classes of solutions produced by our framework. Remember that, by construction, a class is a cluster of highly similar solutions. Figure~\ref{fig:single-class-prop-k2-k3-k4} displays the proportions of cases for which there is a single solution class, as a function of the detected imbalance $I(P)$. Note that we do not include the instances for which there is only a \textit{unique} optimal solution, as they were already discussed before. This results in the absence of certain histogram bars in the plot.

For all values of $\ell_0$, it appears that our method always detects a single class for slightly imbalanced graphs, and that the number of classes increases with the imbalance. There is an exception for $\ell_0=2$ though, as the proportion of single class cases increases again for $I(P)=[0.30,0.35[$ in small graphs. This is surprising, as it corresponds to the largest number of solutions (see Subfigure~\ref{fig:nb-sol-realImb-k0=2}), but could be explained by the concept of elongated class developed next. Overall, single class instances represent $66\%$ of the cases for $\ell_0=2$, $68\%$ for $\ell_0=3$ and $74\%$ for $\ell_0=4$. 

To summarize our findings up to now, graphs with small imbalance tend to lead to a unique solution, and even when there are several, these tend to constitute a single class ($98\%$ of the cases with a detected imbalance $I(P)\in [0.05,0.15[$ and $\ell_0 \in \{2,3,4\}$). Again, since certain real-world networks exhibit such a small imbalance, this seems to indicate that it is not necessary to explore further the solution space in this case. However, this statement does not hold in general, as many of our generated graphs do have a higher imbalance. Moreover, the results shown in Figure~\ref{fig:single-class-prop-k2-k3-k4} do not reflect the inner structure of the detected classes, which can take an ``elongated'' shape. If such a class is indeed a dense group of \textit{locally} similar solutions, its most extreme members are nevertheless quite different. Our core part analysis is meant to study this aspect, and more generally to assess class quality.
\subsection{Analysis of the Core Parts}
\label{subsec:ResultsCoreParts}
We now turn to the characterization and comparison of the classes, through the analysis of their core parts. As a reminder, the core part corresponds to the maximal subset of vertices that always belong to the same modules over all solutions constituting the class. We express the size of a core part in terms of proportion of the graph order $n$ (number of vertices). Our assumption is that, for a class to be considered as cohesive, its core part should be large enough. On the contrary, if the classes are clearly separated, the \textit{overall} core part (processed over all solutions) should be small.

\begin{figure}[!htb]
\captionsetup{width=0.9\textwidth}
    \subfloat{
        \centerline{\includegraphics[height=0.325\textheight,clip=true]{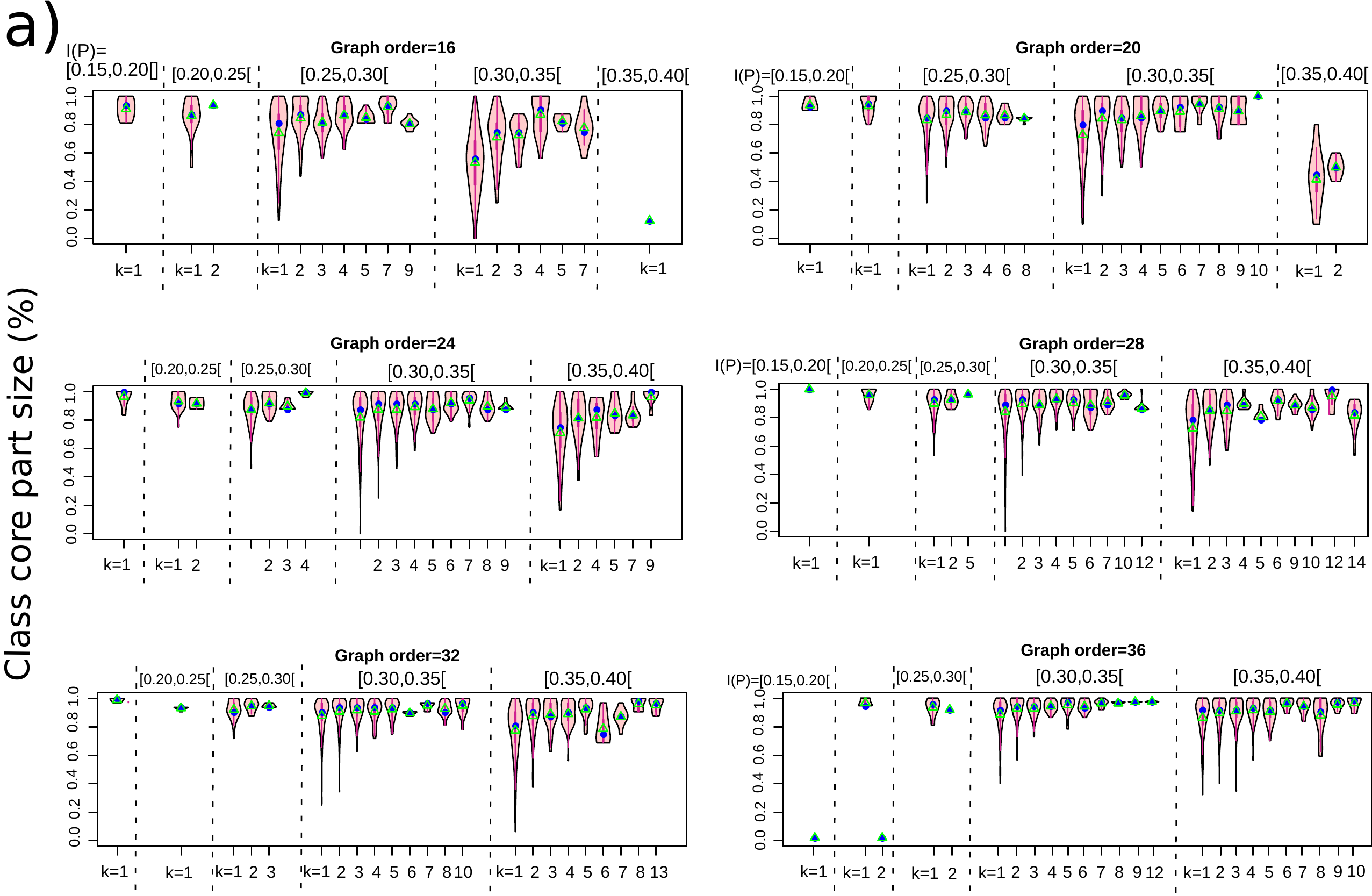}
        \label{fig:class-core-part-size-k0=2}}
    }\\
    \centerline{\subfloat{
        \includegraphics[height=0.250\textheight,clip=true]{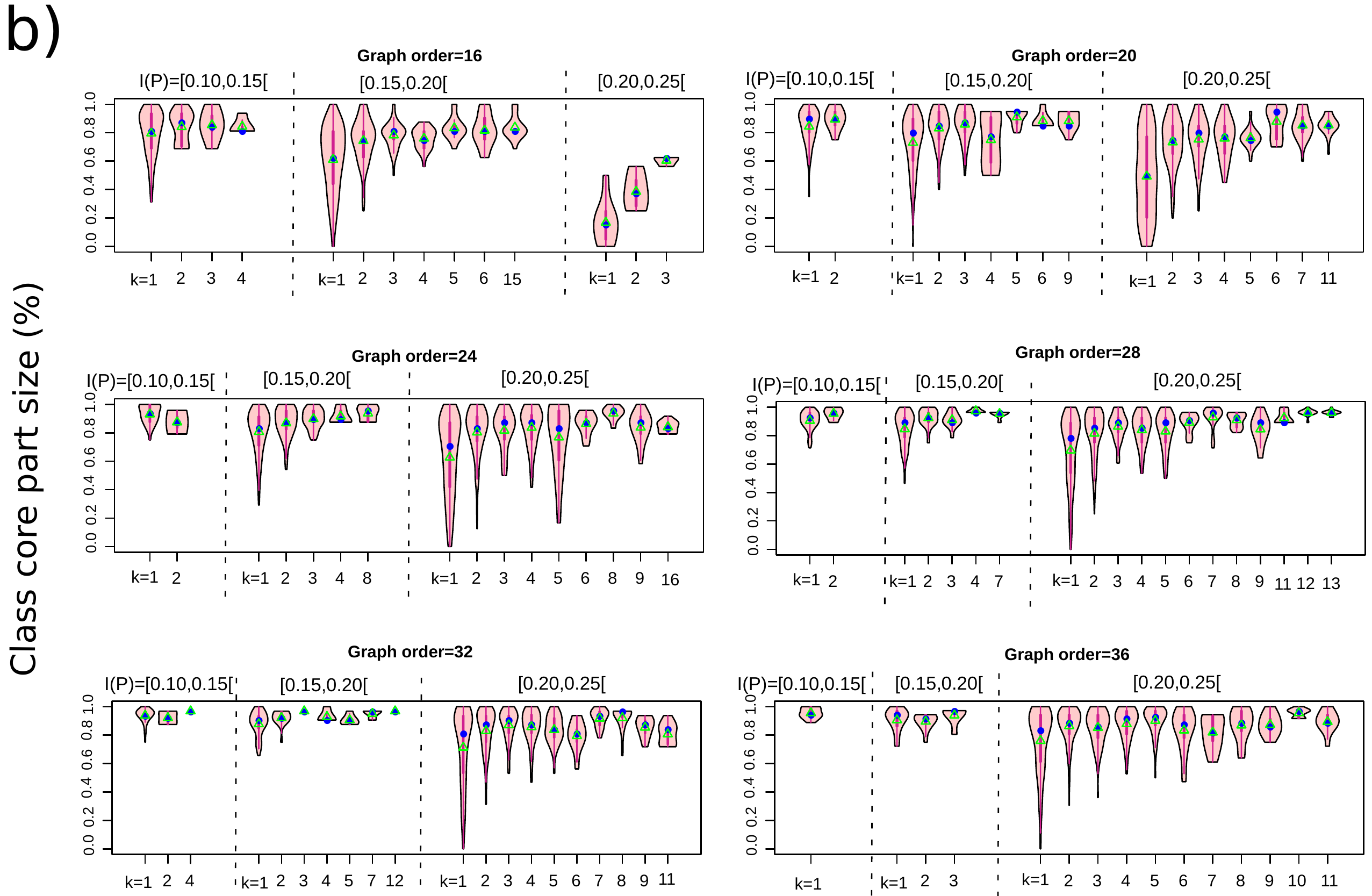}
        \label{fig:class-core-part-size-k0=3}
    }
    \subfloat{
        \includegraphics[height=0.250\textheight,clip=true]{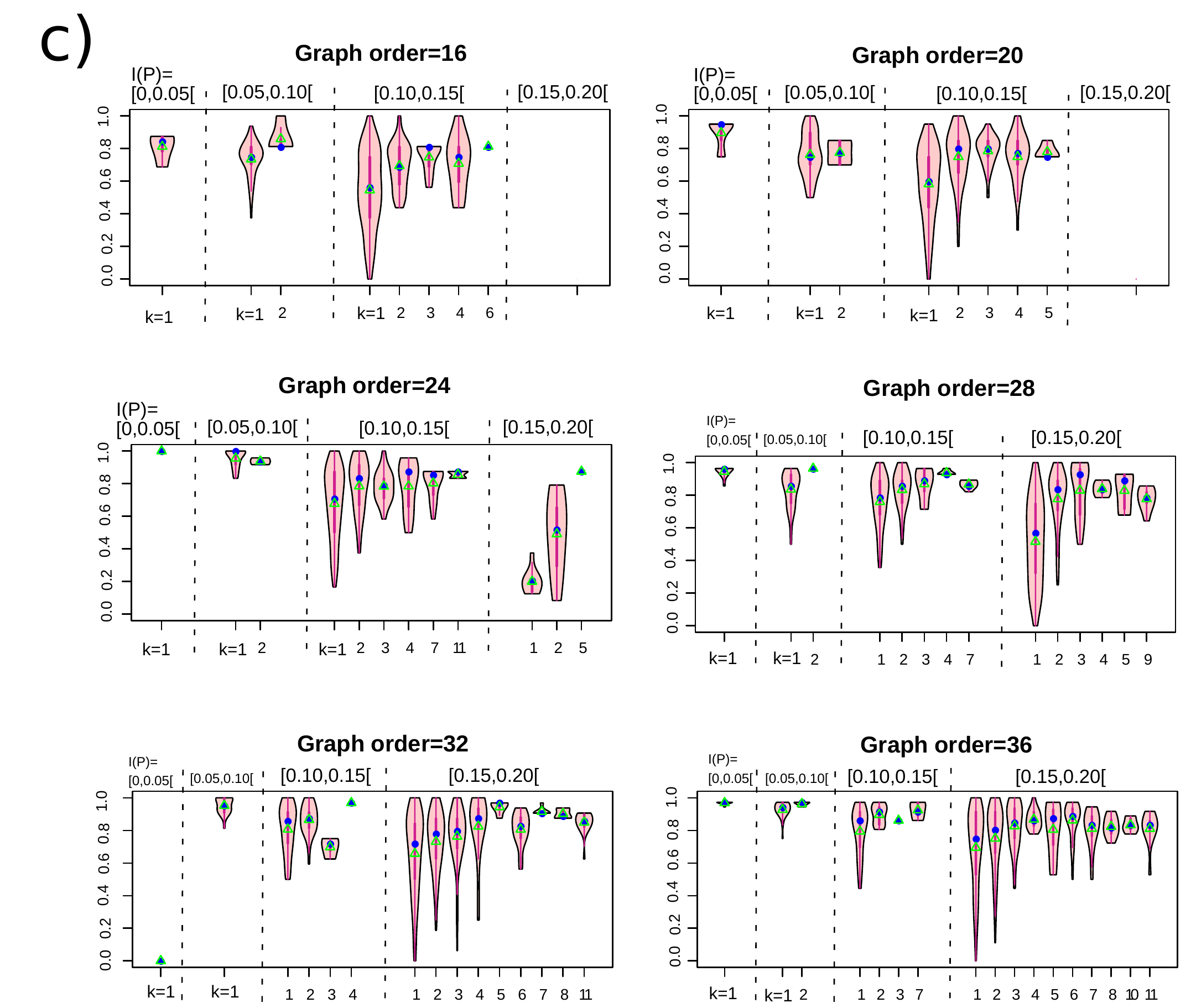}
        \label{fig:class-core-part-size-k0=4}
    }}
    \caption{Proportion of the graph covered by the class core parts, as a function of the detected imbalance $I(P)$ and number of classes $k$, (a) for $\ell_0=2$, (b) for $\ell_0=3$ and (c) for $\ell_0=4$. Notice that an $x$-axis value may be empty if the parameter set is not defined or no data is available. Plots available at \href{https://doi.org/10.6084/m9.figshare.8233340}{10.6084/m9.figshare.8233340} under CC-BY license.}
    \label{fig:class-core-part-size-k2-k3-k4}
\end{figure}

Figure~\ref{fig:class-core-part-size-k2-k3-k4} shows the distribution of class core part sizes as a function of $k$, the number of solution classes (bottom $x$-axis). In addition, the values are grouped using the detected imbalance $I(P)$ (top $x$-axis of each plot). Like before, these plots do not show cases with only a unique solution.

Our first observation is that the core part size seems to increase with the number of classes $k$, at least until it reaches a plateau. This means that the classes are more and more cohesive internally. Moreover, the dispersion also decreases when $k$ increases. In the single-class case, the core part size can be extremely small, close to zero. This indicates that, in certain cases, the cluster analysis is not conclusive: the Silhouette score is too low (below the threshold) to conclude there are several classes, but the single class is not cohesive, and contains some sensibly different solutions.
For a specific real-world application, one would need to manually consider this situation.


Let us now conclude this section related to synthetic networks. We empirically identified four different types of solution spaces. In the first, which tends to happen in only slightly imbalanced graphs, there is only one optimal solution. The second type corresponds to the case where there are multiple solutions distributed over several distinct and cohesive classes. This tends to happen for larger imbalance values. In the third type, we have a single class containing multiple solutions that are very similar, resulting in a large core. A small core means that this class is not cohesive, and corresponds to the fourth type. This typology shows that the answers to our initial questions are multiple and depend on the considered graph. 
Our work highlights the necessity to develop a method allowing to handle these different cases.




\subsection{Real-World Example}
\label{subsec:ResultsSyria}

\begin{figure}[!ht]
\captionsetup{width=0.9\textwidth}
    \begin{center}
     \includegraphics[width=0.75\textwidth]{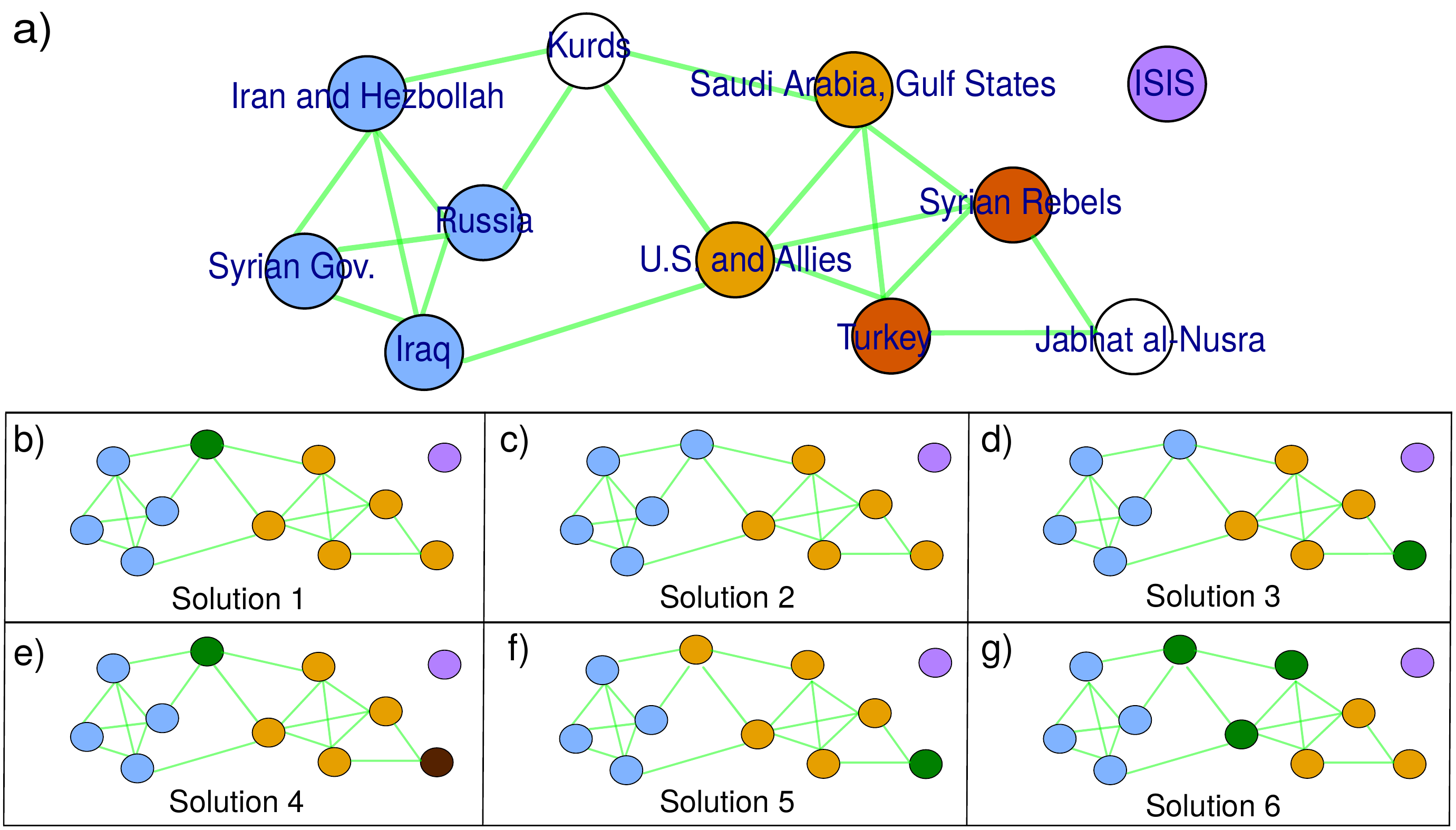}
    \end{center}
    \caption{\textit{a)} Signed graph representing the Syria conflict in 2015. \textit{b}--\textit{g)} show optimal solutions. The graph is complete, but for the sake of clarity, only positive edges are shown (the missing ones thus represent negative edges). Colored vertices constitute the core part in \textit{a)} (non-core vertices are white), and the module assignments for the other graphs. Figure available at \href{https://doi.org/10.6084/m9.figshare.8233340}{10.6084/m9.figshare.8233340} under CC-BY license.}
    \label{fig:real-case-example}
\end{figure}

To show the relevance of the questions at the origin of our work, as well as the usefulness of our method, we further analyze a small real-world graph representing the relations between the main actors of the ongoing Syrian conflict. We choose this dataset because of its size, which eases the interpretation of the obtained CC solutions, but also because it depicts a very interesting situation, as the affiliations of the involved parties are multiple: they are positioned relative to terrorist group ISIS, the Syrian government, and various other geopolitical interests. 

Our source is the 2015 press article \textit{A Guide to Who Is Fighting Whom in Syria} published by Keating \& Kirk in the online news magazine Slate\footnote{\scriptsize \url{www.slate.com/blogs/the_slatest/2015/10/06/syrian_conflict_relationships_explained.html}}, which aims at depicting the Syrian situation as it was in 2015. This is a journalistic work, not an academic work, and it contains certain simplifications, for instance several distinct entities are collapsed together to ease understanding. However, we deem it sufficient in our case, as we are not Political Science or International Relations scholars ourselves, and do not intend at making a thorough political analysis of the situation, but simply use the article content for illustration purposes. 

The article is constituted of a chart and its textual description. The chart is an update of the so-called \textit{Middle East Friendship Chart}, which lists the actors of the conflict as well as the nature of their interactions: \textit{enmity}, \textit{friendship}, or \textit{complicated}. The latter do not correspond to neutral relationships, but rather to undetermined ones, corresponding to a mix of hostile and friendly interactions. The associated text discusses this chart and explains the undetermined relationships. To build a signed graph based on this article, we interpret the chart as the adjacency matrix of a signed graph, representing the operating forces by vertices and their enmity and friendship relationships with negative and positive edges, respectively. Moreover, to keep the network complete and for the sake of illustration, we resolve undetermined relationships into hostile or friendly ones, by leveraging the analysis carried in the text.
This results in the 
network presented in Figure~\ref{fig:real-case-example}.

We apply our framework to the Syria graph, like we did with the synthetic networks. Solving the CC problem yields 6 optimal solutions, each containing 7 frustrated edges (i.e. 12\% of the edges). We describe and discuss each of them for the sake of completeness. In all the solutions, the actors are positioned relative to terrorist group ISIS, which is always detected as a single-vertex module. On top of that, the first to fifth solutions are bipolar: they contrast a pro- (Syrian government, Russia, Iran-Hezbollah, and Iraq) 
and an anti-Syrian government modules, whereas the module assignments of the rest of the actors (Kurds, Jabhat al-Nusra, and ISIS) are not consistent.
The sixth (and last) solution is tripolar: the anti-Syrian government module is split in two: a pro-Kurds module (Kurds, U.S. and Allies, and Saudi Arabia-Gulf States) and an anti-Kurds one (Jabhat al-Nusra, Syrian rebels, and Turkey). Overall, the solution space shows that even for solutions differing only in the assignment of one vertex, the interpretation may change substantially (e.g. whether Kurds forms an alliance with the Syrian government or not). This confirms the necessity to explore the solution space.

When performing the cluster analysis over the space of optimal solutions, we obtain a maximal Silhouette score of $0.315$ for $k=2$, which corresponds to the two types of solutions identified manually above (1st--5th vs. 6th). Although this value is below Kaufman \& Rousseeuw's $0.51$ threshold (cf. Section~\ref{subsec:PerformingClustering}), this clustering makes a lot of sense here, and shows that this threshold is not necessarily always relevant. The overall core part for these classes
is represented by the vertex colors in Subfigures~\ref{fig:real-case-example}a: each color corresponds to a maximal group of vertices always assigned to the same module over all solutions. This highlights how core parts can be used to interpret the differences/similarities between the solution classes. Indeed, the figure reflects common knowledge regarding the geopolitical situation: the tight relationship between the USA and Saudi Arabia, Turkey supporting the Free Syrian Army to create a \textit{buffer zone} in northern Syria from Kurds, and the disagreement between the USA and Turkey regarding Kurds. Interestingly, ISIS is a core vertex, as it is never placed with other vertices.

\section{Conclusion}
\label{sec:Conclusion}
In this work, we empirically studied the space of optimal solutions for the CC problem. We randomly generated a collection of complete synthetic networks and identified all their optimal solutions to obtain their respective solution spaces. We then analyzed these spaces through our cluster analysis-based framework. Our main finding is the identification of 4 different situations: 1) unique solution; 2) single class of similar solutions; 3) several classes of similar solutions; 4) multiple solutions without a clear clustering structure. We also showed that slightly imbalanced networks ($I(P)m<0.15$) tend to be of type 1 or 2, whereas a higher imbalance leads to more solutions, and often several classes. Finally, we illustrated the usefulness of our framework on a small real-world network.


Our work can be extended in several ways. First, the most straightforward perspectives are to apply our framework to incomplete and/or weighted signed graphs; and to consider quasi-optimal solutions for large graphs, following the example of Good \textit{et al.}~\cite{Good2010} with unsigned networks. 
Second, certain steps of our pipeline could be improved. The detection of single-class cases is not satisfying, as it can lead to undetermined situations. Maybe certain solution spaces do not have a crisp clustering structure, in which case a fuzzy clustering method could be more appropriate. 
Third, we plan to do a thorough investigation in order to determine whether core vertices possess certain specific topological properties compared to other vertices. 
Fourth, our results could be used to improve the search for optimal solutions. From a practical perspective, it is not possible to exhaustively enumerate all optimal solutions. 
However, we could leverage the concept of class of similar solutions to design algorithms able to exploit a known optimal solution and find optimal solutions belonging to other classes. 
Such an approach would produce a set of diverse optimal solutions offering a better summary of the whole solution space than the traditional single optimal solution discussed in this paper. 
Fifth, we could work directly on the CC problem itself to reduce the number of optimal solutions. This can be done by optimizing a different imbalance measure (e.g. cycle-~\cite{Cartwright1956} or walk-based measures~\cite{Estrada2019}), capable to discriminate between partitions otherwise considered optimal by the classic imbalance used in this article. It is also possible to add extra constraints in the problem formulation, e.g. by requiring modules to be internally connected in a stronger way (similarly to~\cite{Benati2017} for the clique partitioning problem).

\section*{Acknowledgments}
This research benefited from the support of Agorantic research federation (FR 3621), as well as the FMJH (Jacques Hadamard Mathematics Foundation) through PGMO (Gaspard Monge Program for Optimisation and operational research), and from the support to this program from EDF, Thales, Orange and Criteo.

\phantomsection\addcontentsline{toc}{section}{References}
\printbibliography
%

\appendix
\section{ILP Model for CC on Weighted Signed Graphs}
\label{apx:Model}
The CC problem can be modeled with ILP, as proposed by Demaine \textit{et al.}~\cite{Demaine2006} for the \textit{Uncapacitated Clustering} problem~\cite{Mehrotra1998}. We include the model here for the sake of completeness, and our source code is publicly available\footnote{\url{https://github.com/CompNet/ExCC}}. Note that the model can handle weighted graphs, but we use it on unweighted ones in the context of this article.

For an undirected signed graph, let $E^-$ and $E^+$ the sets of negative and positive edges in the signed graph, respectively. Moreover, a signed graph is weighted if there is a function $w : E^- \cup E^+ \rightarrow {\rm I\!R}^+$.

For all vertices $i, j \in V: i < j$, a binary set is first defined to describe pairs of vertices that are in the same module
\begin{equation*}
    x_{ij} = 
    \begin{cases} 
        1, & \mbox{if $i$ and $j$ are in a same module,} \\
        0, & \mbox{otherwise}.
    \end{cases}
\end{equation*}

Then, the ILP formulation of the CC problem for weighted signed graphs is written as follows:
\begin{align}
    \text{Min} \quad & \sum_{i,j \in V:ij \in E^-} w_{ij} x_{ij} + \sum_{i,j \in V:ij \in E^+} w_{ij} (1- x_{ij}) \label{eq:cc_obj} \\
    \text{s.t.} \quad  & x_{ij} + x_{jr} - x_{ir} \leq 1 ,\quad \forall i<j<r \in V \label{eq:S1_cc2}
    \\
    & x_{ij} - x_{jr} + x_{ir} \leq 1 ,\quad \forall i<j<r \in V \label{eq:S1_cc3}
    \\ 
    & -x_{ij} + x_{jr} + x_{ir} \leq 1 ,\quad \forall i<j<r \in V \label{eq:S1_cc4}
    \\
    & x_{ij} \in \{0,1\} ,\quad \forall i,j \in V. \label{eq:S1_cc5}
\end{align}

Our goal is to minimize the objective function (\ref{eq:cc_obj}) by finding a valid assignment for the set of $x_{ij}$ variables. An assignment is valid (corresponds to a partition) if $x_{ij}$ is either 0 or 1 (\ref{eq:S1_cc5}); and the $x_{ij}$ variables satisfy the triangle inequalities (\ref{eq:S1_cc2}, \ref{eq:S1_cc3}, \ref{eq:S1_cc4}). For instance, the triangle inequality in (\ref{eq:S1_cc2}) says that if vertices $i$ and $j$ are in a same module, as well as vertices $j$ and $r$, then vertices $i$ and $r$ are  also in this same module.

As explained in Section~\ref{subsec:EnumOptPartitions}, we further strengthen this ILP model with tight valid inequalities generated during the root relaxation phase through a \textit{cutting plane} approach. We use the 2-partition and the 2-chorded cycle inequalities, whose efficiency is empirically proved by Ales \textit{et al.}~\cite{Ales2016a}.




\end{document}